\documentclass[11pt]{article}
\usepackage[utf8]{inputenc}
\usepackage{amssymb}%
\usepackage{amsthm}%
\usepackage{amsmath}
\usepackage{amsfonts}
\usepackage{dsfont}
\usepackage{stmaryrd}
\usepackage[table,xcdraw]{xcolor}%
\usepackage[caption=false,font=footnotesize,labelfont=sf,textfont=sf]{subfig}
\usepackage[font=footnotesize]{caption}
\usepackage{graphicx}
\usepackage{authblk}
\usepackage{natbib}
\usepackage{hyperref}
\hypersetup{%
        colorlinks,
        breaklinks=true,
        plainpages=false,%
        citecolor=blue,
        linkcolor=blue,
        urlcolor=blue,
        bookmarksopen=true,%
        bookmarksnumbered=false,%
        bookmarksdepth=5%
}

\usepackage{fullpage}

\newcommand{\norm}[1]{\left\lVert#1\right\rVert}
\newcommand{\parent}[1]{\left(#1\right)}
\newcommand{\llbrack}[1]{\left\llbracket#1\right\rrbracket}

\renewcommand{\brack}[1]{\left[#1\right]}
\renewcommand{\brace}[1]{\left\{#1\right\}}
\renewcommand{\det}[1]{\lvert#1\rvert}

\newcommand{\argmin}[1]{\underset{#1}{\text{argmin}} \,}
\newcommand{\argmax}[1]{\underset{#1}{\text{argmax}} \,}
\renewcommand{\min}[1]{\underset{#1}{\text{min}} \,}
\renewcommand{\max}[1]{\underset{#1}{\text{max}} \,}

\renewcommand{\P}{\mathbb{P}}
\newcommand{\R}{\mathbb{R}}
\newcommand{\Rcal}{\mathcal{R}}

\newcommand{\N}{\mathcal{N}}
\newcommand{\bN}{\mathbb{N}}
\newcommand{\F}{\mathcal{F}}

\newcommand{\D}{\mathcal{D}}
\newcommand{\indic}{\mathds{1}}
\newcommand{\eqname}{Eq.~}
\newcommand{\sectionname}{Section}

\newcommand{\muchless}{\! \ll \!}

\newcommand{\pr}{\text{Pr}}
\newcommand{\muf}{\mu_{2}}
\newcommand{\mut}{\mu_{1}}
\newcommand{\St}{\Sigma_1}
\newcommand{\Sf}{\Sigma_2}
\newcommand{\mufh}{\hat{\mu}_{2}}
\newcommand{\muth}{\hat{\mu}_{1}}
\newcommand{\Sth}{\hat{\Sigma}_1}
\newcommand{\Sfh}{\hat{\Sigma}_2}
\newcommand{\ny}{\left\{}
\newcommand{\zr}{\right\}}
\newcommand{\mh}{\hat{m}_1}
\newcommand{\Sh}{\hat{S}_1}

\providecommand{\keywords}[1]
{
  \small	
  \textbf{\textit{Keywords---}} #1
}

\newtheorem{theorem}{Theorem}
\newtheorem{proposition}[theorem]{Proposition}%
\newtheorem{corollary}{Corollary}

\newtheorem{example}{Example}%
\newtheorem{remark}{Remark}%

\begin{document}

\title{On unsupervised projections and second order signals}


\author[1]{Thomas Lartigue}
\author[1,2]{Sach Mukherjee}
\affil[1]{Statistics and Machine Learning, German Center for Neurodegenerative Diseases, Bonn, Germany}
\affil[2]{MRC Biostatistics Unit, University of Cambridge, UK}
\date{}                     
\setcounter{Maxaffil}{0}
\renewcommand\Affilfont{\itshape\small}

\maketitle


\begin{abstract}
    Linear projections are widely used in the analysis of high-dimensional data. In unsupervised settings where the data harbour latent classes/clusters, the question of whether class discriminatory signals are retained under projection is crucial. In the case of mean differences between classes, this question has been well studied. However, in many contemporary applications, notably in biomedicine, group differences at the level of covariance or graphical model structure are important. Motivated by such applications, in this paper we ask whether linear projections can preserve differences in second order structure between latent groups. We focus on unsupervised projections, which can be computed without knowledge of class labels. We discuss a simple theoretical framework to study the behaviour of such projections which we use to inform an analysis via quasi-exhaustive enumeration. This allows us to consider the performance, over more than a hundred thousand sets of data-generating population parameters, of two popular projections, namely random projections (RP) and Principal Component Analysis (PCA). Across this broad range of regimes, PCA turns out to be more effective at retaining second order signals than RP and is often even competitive with supervised projection. We complement these results with fully empirical experiments showing 0-1 loss using simulated and real data. We study also the effect of projection dimension, drawing attention to a bias-variance trade-off in this respect. Our results show that PCA can indeed be a suitable first-step for unsupervised analysis, including in cases where differential covariance or graphical model structure are of interest.
\end{abstract}




\section{Introduction}
Linear projections are widely used to reduce data dimension
within high-dimensional analysis workflows,  motivated by both statistical and computational considerations. When data span multiple classes/clusters, an important question is whether projections retain signals that discriminate between the classes. 
For mean signals -- i.e. differences in class-specific means --
a number of classical results are available to understand the effect of projection (e.g. the Johnson-Lindenstrauss lemma supporting the use of random projections as a signal-retaining transformation \citep{dasgupta2003elementary}). However, in many contemporary applications, 
differential {\it covariance} structure between classes/clusters
plays an important role and when 
mean signals are not very strong, such second order signals may be essential in distinguishing between classes/clusters.
This issue -- of subtle differences between classes leading to mean differences that are relatively small in magnitude -- is relevant in biomedical applications, where differences in covariance structure, graphical models or networks are often important. Examples include differences in gene or protein networks \citep{ideker2012differential,stadler2017molecular} and in patterns of neural activity 
\citep{varoquaux2010brain,smith2011network}.


In the general setting where both mean and covariance information may be relevant, a standard approach to unsupervised learning is to use mixture models 
with component distributions suitable to model the underlying covariance structures \citep{scrucca2016mclust}. 
Flexible/full covariance mixtures are well studied in low dimensions, but remain challenging in high dimensions, due to both statistical and computational factors \citep{stadler2017molecular}. For example, EM-type algorithms for the estimation of mixtures of high-dimensional graphical models \citep{stadler2017molecular} must cope with multiple calls to the graphical model estimators, typically leading to a heavy computational burden in high dimensional problems. 

Linear projections have been widely studied for unsupervised learning in high-dimensional settings 
\citep[e.g.][]{li2006very,jin2016influential}.
One natural strategy is to project $p$-dimensional data into a $q$-dimensional space, with $q \! \muchless \! p$, and then perform mixture modelling in the lower dimensional space (rather than in the original, ambient space). If relevant differences between latent groups are retained under projection, approaches of this kind are attractive because the main iterative computations and statistical estimation steps are carried out in $q$ rather than $p$ dimensions. 
An example of this type appears in \citet{taschler2019model}, who show how such an approach can be used to not only learn cluster labels, but also {\it high-dimensional} model parameters. 
More generally,  workflows of this kind -- involving projection followed by clustering or mixture modelling -- are common in practical data analysis in many fields.

While simple and attractive, the  strategy of projection followed by mixture modelling in the lower dimensional space is subject to the clear limitation that if the projection does {\it not} retain the required signal (we make these notions precise below), then no estimator in the second stage, however sophisticated, will be able to effectively model the latent structure. That is, signal retention (suitably defined) is a necessary condition for effective learning.

\bigskip

\noindent
{\it Problem statement.} Thus, motivated by  developments at the intersection between unsupervised learning
and high-dimensional  applications (particularly in biomedicine), in this paper we set out to study whether linear projections can retain second order, covariance signals. 
To limit scope, we focus on the case of class-specific Gaussian distributions in the high-dimensional space, although many elements of the conceptual framing and analysis will be relevant for other distributions (our experiments include also real non-Gaussian data). 
Consider data of the form $$(x_i, z_i)_{i=1,\ldots,n}, \, x_i \in \mathbb{R}^p, \, z_i \in \{ 1, \ldots, K \},$$ where the $x$'s are data vectors and the $z$'s are associated discrete labels. In the unsupervised problems that motivate this work, the $z$'s are latent and not available to the learner.
The data follow class-specific distributions, i.e.  $(x_i \vert z_i = k) \! \sim \! p_k$ and it is differences in the densities $p_k$ that permit learning of latent class labels. 
In the conditionally Gaussian case we have $(x_i \vert z_i = k) \! \sim \! \N(\mu_k,\Sigma_k)$, where $\N$ denotes a multivariate Gaussian distribution with mean $\mu_k$ and $p \times p$ covariance matrix $\Sigma_k$. Here, differences may exist wrt means, covariances or both. We refer to differences between the class-specific means $\mu_1, \ldots, \mu_K$ as the {\it first order} or {\it mean signal} and differences in the covariances $\Sigma_1, \ldots, \Sigma_K$ as the {\it second order} or {\it covariance signal}. A linear projection via a $p \times q$ matrix $W$ (typically with $q \muchless p$) yields low-dimensional data $W^t x_i \! \in \! \mathbb{R}^q$. Unsupervised learning using these reduced dimension data requires that sufficient signal is retained under projection to allow learning of the latent labels $z_i$. 

Our aim is to provide a framework within which to study 
the ability of linear projections to retain this second order signal, i.e. to allow discrimination between classes/clusters with different $p$-dimensional covariances $\Sigma_k$ on the basis of the low-dimensional data $W^t x_i$. We emphasize that we focus on the case of projections that are unsupervised in the sense that the projection step does not have access to any class labels.
To this end, we discuss a risk-based framework within which to consider the behaviour of projections and use this framework to address specific questions, including the behaviour of random projections and PCA with respect to second order signals and 
the choice of projection dimension $q$.

%
%

\bigskip

\noindent
{\it Summary of contributions.}
The main contributions of this paper are:
\begin{itemize}
\item {\it Conceptual framework}. We study the behaviour of projections through the lens of 0-1 loss and in particular consider the Bayes risk under projection and bounds thereupon as a measure of signal retention. 
\item {\it Comparison of PCA and RP through quasi-exhaustive enumeration}.
We use a closed-form distribution-overlap metric to conveniently assess class-separability within any embedding space. With it, we are able to explore three families of covariance matrices through quasi-exhaustive enumeration, a task that would be infeasible via brute-force 
Monte Carlo estimation of the Bayes risk. These results demonstrate that PCA projection clearly outperforms RP. We go on to study this phenomenon via a 0-1 loss experimental analysis on finite data. 
\item {\it Choice of projection dimension $q$}. We draw attention to the importance of the projection dimension $q$ in governing a type of bias-variance trade off. The analysis underlines the importance of setting $q$ appropriately and in particular shows that the choice of $q{=}K$ (where $K$ is the number of classes), as suggested by the well-known correspondence between K-means and PCA \citep{ding2004k}, may be far from optimal.

\end{itemize}

\bigskip

\noindent
{\it Key observations.} Our analysis and results lead to several interesting observations. Three key observations are: (i) PCA projection is able to preserve second order signal even with {\it no} mean difference, confirming an empirical observation of \citet{taschler2019model}.
(ii) PCA projection is more effective than RP at this task. (iii) PCA projection remains effective
for class recovery at moderate sample sizes. (Interestingly, this latter point does not hold for the 
direct empirical analogue of the 
theoretically optimal projection, due to estimation errors wrt high-dimensional covariance matrices).

Given the low computational cost of PCA when at most one of the dimension $p$ or the sample size $n$ is very large, the confirmation that cluster identification is still very much possible in PCA-defined embedding spaces supports the use of PCA projection as 
an element of computationally-efficient unsupervised learning even in settings where second order structure is relevant.

\bigskip

\noindent
{\it Organization.}
The remainder of this paper is organized as follows. 
In \sectionname~\ref{sect:litterature} we provide an in-depth review of the literature relevant to supervised and unsupervised classification within projected subspaces. In \sectionname~\ref{sect:theory} we review basic notions needed for a 0-1 loss analysis
including in particular the computationally convenient Bhattacharyya bound on the Bayes risk \citep[see][]{bhattacharyya1943measure, chernoff1952measure}.
Then, using these notions, we explore the effects of projection, including risk behaviour and the impact of the projection dimension $q$.
With these basic ideas established, we study in \sectionname~\ref{sect:PCA_theory} the theoretical 0-1 loss under PCA projection in the no mean-difference two-class Gaussian case. We show that without assumption on the within-class covariance matrices non-trivial bounds on the PCA projection risk are not possible nor on regret with regards to the risk-optimal projection. As a consequence, we explore several specific families of matrices in \sectionname~\ref{sect:semi_theoretical}. Thanks to the computational convenience afforded by the Bhattacharyya bound, 
we then carry out a quasi-exhaustive numerical exploration of class separation behaviour under a large number of data-generating regimes.
In \sectionname~\ref{sect:classifier_error}, we then complement these quasi-exhaustive results with fully empirical results evaluating 0-1 loss achieved through analysis of (finite) real data. Finally, In \sectionname~\ref{sect:comparison_optimal}, we compare PCA projection and RP to the Bhattacharyya-optimal projection on both simulated and real data.

\section{Literature review}\label{sect:litterature}
In this section we review the extensive literature on linear projections including the interface with supervised and unsupervised learning of class/cluster labels and the case where covariance structure -- i.e. the second order signal -- plays a role.


\subsection{Model-based classifiers}
To analyse multi-class data, model-based methods such as likelihood ratio classifiers \citep{pickles1985introduction, severini2000likelihood} are popular and interpretable tools to separate observations into different classes.
In supervised and unsupervised scenarios, such models are naturally estimated through maximum likelihood estimation of a hierarchical or mixture model respectively \citep{everitt1981finite}. 

The Bayes risk (i.e. the risk achieved by a Bayes' classifier based on the population parameters) is a reference for the fundamental class-separability in a given space. Indeed, if the {Bayes classifier} performs poorly, so will all model-based classifiers, whether trained on a labelled or unlabelled dataset. 
In the two-class Gaussian case, the \textit{Bhattacharyya distance} \citep{bhattacharyya1943measure}, a classical measure of divergence between two distributions, provides an upper bound on the {Bayes risk}. This \textit{Bhattacharyya bound} \citep[see for instance][]{fukunaga1972instruction}, represents an attractive, computationally efficient substitute to the Bayes risk
and this has been optimised to define embedding subspaces for classification 
\citep{kullback1997information, kadota1967best, henderson1969comments, guorong1996bhattacharyya, xuan2006feature}. 
Bhattacharyya-optimal embeddings are available in closed form in the zero mean-difference case, \citep[see][]{kadota1967best, henderson1969comments, fukunaga1972instruction}. However, their computation requires knowledge of the class labels and hence cannot be 
performed in the purely unsupervised case. 


\subsection{Unsupervised projection and clustering.} 
There is a rich literature on projections, including RP and PCA,  for the unsupervised setting.
Much effort has been made to find mean separability conditions that guarantee a high probability of cluster identification in the embedding space. Under a mixture of spherical Gaussian model assumption, \cite{dasgupta1999learning} proved that above a certain level of mean separability
classification on randomly projected data is effective. 
Subsequent works have extended and improved this analysis, including \citep{dasgupta2000two,sanjeev2001learning,vempala2002spectral,kannan2005spectral, achlioptas2005spectral,brubaker2008isotropic}.
When Gaussian components are axis-aligned, \cite{feldman2006pac} have shown that the mixture can be learnt without any separation assumption. 
See \cite{kannan2005spectral} or \cite{feldman2006pac} for more detailed reviews of each of these methods. 

%

Clustering with unsupervised embeddings is a vast topic, and a wide range of  questions have been addressed. Works such as \cite{fradkin2003experiments} and \cite{deegalla2006reducing} have empirically compared PCA projection and random projection on real data sets. Both concluded that PCA is better but RP is faster and easier to compute. \cite{kim2003efficient} and  \cite{xu2014multimode} have been learning Gaussian Mixture models with PCA projection. \citet{ailon2009fast} define a new type of sparse random embedding, in the footsteps of \cite{achlioptas2003database}, that preserves well the mean difference. \citet{houdard2018high} learn high dimensional Gaussian mixtures by assuming that the real signal lies in a lower latent dimension. 
\cite{hertrich2020pca} propose the PCA-GMM model, where the PCA dimension reduction appears as a parameter of the model and is optimised within the EM.

Of particular interest to us, \cite{taschler2019model} show empirically that, even with {\it no} mean difference, model-based clustering within a projected space can work well, in particular with the PCA projection. To the best of our knowledge, there is no clear explanation for this observation in the unsupervised literature. Hence, we consider the supervised literature for tools to 
study this behaviour.

\subsection{Supervised projections and classification} 
Supervised results are relevant to understanding unsupervised analysis.
A supervised projection defined to maximise class separability in the embedding space 
gives information on what a good projection may look like.
Further, a classifier trained 
in a supervised sense on an unsupervised projection provides a ``label oracle" best case scenario; when 
the supervised 0-1 loss is poor, then clustering methods are expected to fail as well.
One branch of the literature does not try to learn projections from the labels, and focuses on classifiers trained within RP subspaces. Works such as \cite{durrant2010compressed}, \cite{durrant2013sharp}, \cite{kaban2020structure}, \cite{kaban2020sufficient} or \cite{reeve2021statistical} 
obtain classification error bounds for learning with randomly projected data. 
Among other things, such bounds provide insight into some general issues of learning under projection, e.g. the way in which the embedding dimension $q$ appears implicitly in the bounds of \cite{reboredo2014compressive}, \cite{durrant2015random} and \cite{reeve2021statistical} 
points to potential trade-offs (that we make more explicit below).
Other notable works include \cite{cannings2017random} and \cite{zhang2019experiments} which study ensembles of random projections. 

When it comes to projections learned from data, several works use Fisher Discriminant Analysis (FDA) \citep{fisher1936use, fukunaga1972instruction} to define good projection directions. This is a classical and very rich branch of the literature \citep[see e.g.][]{hastie1996discriminant,mika1999fisher,sugiyama2007dimensionality}. However, all these works seek to maximise some form of Fisher's Linear Discriminant which does not have a nontrivial form when the mean difference is zero and hence cannot be applied to subspace selection in this case. 

Hence, we look to the projection defined from optimisation of the Bhattacharyya distance \citep{bhattacharyya1943measure}. The Bayes risk is upper bounded by the Bhattacharyya bound, which has an analytical formula. 
Additionally, the Bhattacharyya distance takes into account the covariance-difference, and has a nontrivial form when the mean-difference is null. 
A branch of the literature looks at subspaces maximising the Bhattacharyya distance 
including classical work on the best direction when $q=1$ \citep{kullback1997information} and for $q>1$ in the equal covariance \citep{van2004detection} or equal mean \citep{kadota1967best} cases (however, there is no analytical result for the optimal subspace in the general case). Algorithmic solutions for approximately optimal sub-spaces have been proposed for the case where either the mean difference of the covariance difference is dominant over the other \citep{henderson1969comments, fukunaga1972instruction}. Later works \citep{guorong1996bhattacharyya, choi2003feature, xuan2006feature}, have proposed algorithms to find sub-spaces in the general case, with no assumption of dominance.


\section{Theoretical elements}\label{sect:theory}
In this section, we recall basic elements of model-based classification theory, such as the \textit{Bhattacharyya bound} on the \textit{Bayes risk}. Then, we study the new stakes emerging from the addition of the projection step, such as the choice of the projection methodology and of the target embedding space size $q$. In particular, we bring to light the trade-off between loss of information and loss of regularity when setting $q$.
Although we are motivated by the unsupervised case, in many places below we present key ideas from a supervised point of view for simplicity of exposition. The relevance of the Bayes risk is twofold. First, as a fundamental measure of separability, since if the Bayes' risk associated with a projection is poor, we cannot hope to perform well with any analysis of the projected data. Second, although the Bayes' risk is an asymptotic/population notion, since we are interested in projections that greatly reduce the dimension such that $q \ll n$, after projection, the regime is typically no longer high-dimensional in a statistical sense. 

\subsection{Background} \label{sect:theory_background}

\subsubsection{Bayes risk} \label{sect:bayes_classifier}
We consider  classification  with $K \! \in \! \bN^* \! \setminus \! \brace{1}$ classes. Let $p \in \bN^*$ be the ambient space dimension and consider $\brace{\D_k}_{k=1}^K$, $K$ distributions over $\R^p$. We denote by $p_k : \R^p \! \rightarrow \! \R_+$ their respective probability density functions (pdf). For $n \in \bN^*$, let $X := (x_i^T)_{i=1}^n \in \R^{n \times p}$ be an observed dataset, where the $x_i$ are independent, and drawn from one of the $\D_k$ with probability $\pi_k$, where $\sum_{k=1}^K \pi_k = 1$. We call $z_i \in \llbrack{1, K}$ the label of the observation $x_i$. If the dataset is unlabelled, the value of $z_i$ has to be guessed from $x_i$. The probability of error of a classifier $h : \R^p \rightarrow \llbrack{1, K}$, is the \textit{risk}: $\Rcal(h) :=  \P(h(x) \neq z)$. When all model parameters are known, then the decision rule with the minimum risk is the \textit{Bayes classifier}:
\begin{equation} \label{eq:bayes_classifier}
z^*(x) := \argmax{k} \brack{\P(z = k \vert x)} = \argmax{k} \brack{\pi_k p_k(x)} \, .
\end{equation}
The associated \textit{Bayes risk} $\epsilon$ verifies:
\begin{equation} \label{eq:bayes_risk}
    \epsilon := \Rcal(z^*) = 1 - \int_x \max{k}\brack{\pi_k p_k(x)} dx  \, . 
\end{equation}
The Bayes risk is a reference for the best case scenario achievable by a model-based classifier. It has a geometrical interpretation, as a measure of the overlap between distributions. The Bayes risk is a fundamental measure of distribution separability.

\subsubsection{Bound on the Bayes risk} \label{sect:bayes_risk_bound}
In the Gaussian two-classes case, that is to say when $K=2$ and $\D_k$ are Gaussian distributions, upper bounds of the Bayes risk can be formulated in analytical form. We recall briefly the formula and proof of the Chernoff bound \citep{chernoff1952measure} following \cite{fukunaga1972instruction}. For $K=2$, the Bayes risk \eqref{eq:bayes_risk} can be expressed as:
\begin{equation*} 
    \epsilon = \int_x \min{k \in \brace{1, 2}} \brack{\pi_k p_k(x)} dx  \, . 
\end{equation*}
For any $a, b \geq 0$ and $0 \leq s \leq 1$, we have $\text{min} \brack{a, b} \leq a^{s} b^{1-s}$. As a result, the following inequality, which is called the \textit{Chernoff bound}, holds:
\begin{equation*} 
    \forall\,  0 \leq s\leq 1, \quad \epsilon \leq \pi_1^{s} \pi_2^{1-s} \int_x  p_1^{s}(x) p_2^{1-s}(x) dx  \, . 
\end{equation*}
With the additional assumption that for $k=1, 2$,  $\D_k = \N_k(\mu_k, \Sigma_k)$, we have for any $0 \leq s\leq 1$:
\begin{equation*} 
     \int_x  p_1^{s}(x) p_2^{1-s}(x) dx = e^{-\delta(s)}  \, ,
\end{equation*}
with $\delta(s)$ the \textit{Chernoff distance}: 
\begin{equation} \label{eq:chernoff_distance}
    \delta(s)=\frac{s(1-s)}{2} (\mu_2-\mu_1)^T (s \Sigma_1 + (1-s)\Sigma_2)^{-1} (\mu_2-\mu_1) + \frac{1}{2}\ln \frac{\det{s \Sigma_1 + (1-s)\Sigma_2}}{\det{\Sigma_1}^s \det{\Sigma_2}^{1-s}} \, . 
\end{equation}
When $\Sigma_1 \neq \Sigma_2$, finding the optimal $s$ for the tighter bound is challenging. If simplicity is favoured over optimality, one can set $s{=}\frac{1}{2}$ and get the \textit{Bhattacharyya distance} \citep{bhattacharyya1943measure}:
\begin{equation} \label{eq:bhattacharyya_distance_general}
     \delta\left(\frac{1}{2}\right) = \frac{1}{2}\ln \frac{\det{\frac{\Sigma_1 + \Sigma_2}{2}}}{\sqrt{\det{\Sigma_1} \det{\Sigma_2}}} + \frac{1}{8} (\mu_2-\mu_1)^T \parent{\frac{\Sigma_1 +\Sigma_2}{2}}^{-1} (\mu_2-\mu_1) \, ,
\end{equation}
and the associated \textit{Bhattacharyya bound}:
\begin{equation} \label{eq:bhattacharyya_bound_general}
    \epsilon \leq (\pi_1\pi_2)^{\frac{1}{2}} e^{-\delta(\frac{1}{2})} =: \epsilon_{\text{BB}} \, . 
\end{equation}
Although this inequality is not always very tight \citep[see][]{fukunaga1972instruction}, it highlights that the upper bound $\epsilon_{\text{BB}}$ is a measure of the distribution overlap with relevance to the classification risk
that is data-free and computationally convenient. We refer to $\epsilon_{\text{BB}}$ as the \textit{Bhattacharyya-overlap} between the two distributions. 

\subsection{Projections} \label{sect:theory_projections}

\subsubsection{Optimal sub-space selection} \label{sect:optimal_subspace}
Let $q \in \bN^*,\,  q \! < \!  p$ be the embedding space size, and $W \in \R^{p\times q}$ be a projection matrix with orthonormal columns. The associated transformation is $x \mapsto W^t x \in \R^q$. The Bayes classifier constructed from the low dimension sketch of the data $W^t x$ is:
\begin{equation} \label{eq:bayes_classifier_embedding}
    z_W^*(x) := \argmax{k} \brack{\P(z = k \vert W^t x)} \, ,
\end{equation}
with associated risk:
\begin{equation} \label{eq:bayes_risk_embedding}
    \epsilon(W) := \P(z_W^*(x) \neq z)\, .
\end{equation}
For instance, in the Mixture of Gaussians (MoG) case, $(x\vert z=k) \sim \N(\mu_k, \Sigma_k)$, the linear projection $W^t x$ follows a known MoG distribution $(W^t x \vert  z = k) \sim \N(W^t \mu_k, W^t \Sigma_k W)$. In this case the embedded parameters $W^t \mu_k$ and $W^t \Sigma_k W$ are the ones used in the definition of the Bayes classifier \eqref{eq:bayes_classifier_embedding}. Regardless of any model assumption, the property of optimality of the Bayes classifier implies that: $\epsilon = \Rcal(z^*) \leq \Rcal(z_W^*) = \epsilon(W)$. Indeed, the loss of information inherent in any projection decreases the class separability. Hence, finding projections that preserve said class separability as much as possible is crucial in the context of classification. For a given embedding space size $q$, the ideal projection $W \in \R^{p\times q}$ would be the one that minimises the embedding Bayes risk $\epsilon(W)$. However, since there is no closed form for the Bayes risk, approaching this problem theoretically is hard. As a result, many authors have worked with the Bhattacharyya bound $\epsilon_{\text{BB}}(W)$ instead, looking for the subspace that minimises the Bhattacharyya-overlap between distributions. As early as 1959, Kullback \citep{kullback1997information} described the best direction for the Bhattacharyya-overlap when $q=1$. The formulas for the optimal subspace of size $q>1$ are well known in the equal covariance case \citep[$\Sigma_1=\Sigma_2$,][]{van2004detection} or equal mean case \citep[$\mu_1=\mu_2$,][]{kadota1967best}. There is no analytical formula for the optimal subspace in the general case. Algorithmic solutions for approximately optimal sub-spaces have been proposed for the case where either the mean difference or the covariance difference is dominant over the other \citep{henderson1969comments, fukunaga1972instruction}. Later works \citep{guorong1996bhattacharyya, choi2003feature, xuan2006feature}, have proposed algorithms to find sub-spaces in the general case, with no assumption of dominance.

Among the more general projection literature, it is well documented that the mean difference signal is preserved by  random linear projection \citep[Johnson–Lindenstrauss lemma,][]{lindenstrauss1984extensions}. The lesser studied question of the fate of the second order (covariance) signal through projection is what motivates our study. Hence we will focus on the zero mean difference case. When $\mu_1{=}\mu_2{=}0$, the Bhattacharyya-overlap \eqref{eq:bhattacharyya_distance_general} reduces to:
\begin{equation} \label{eq:bhattacharyya_distance}
     \delta(\frac{1}{2}) =  \frac{1}{2}\ln \frac{\det{\frac{\Sigma_1 + \Sigma_2}{2}}}{\sqrt{\det{\Sigma_1} \det{\Sigma_2}}} \, ,
\end{equation}
and the Bhattacharyya bound \eqref{eq:bhattacharyya_bound_general} to:
\begin{equation} \label{eq:bhattacharyya_bound}
    \epsilon \leq \epsilon_{\text{BB}} = (\pi_1\pi_2)^{\frac{1}{2}} \parent{\frac{\det{\frac{\Sigma_1 + \Sigma_2}{2}}}{\sqrt{\det{\Sigma_1} \det{\Sigma_2}}}}^{-\frac{1}{2}} \, . 
\end{equation}
Let $\phi_1, ..., \phi_p$ be the eigenvectors of $\Sigma_1^{-1} \Sigma_2$ and $\lambda_1, ..., \lambda_p$ the associated eigenvalues. Assume that they are ordered such that $\lambda_1 + \frac{1}{\lambda_1} > ... > \lambda_p + \frac{1}{\lambda_p}$. Then, with $q$ fixed, the projection $W \in \R^{p \times q}$ that minimises the embedded Bhattacharyya-overlap,
\begin{equation} \label{eq:bhattacharyya_bound_embbeding}
    \epsilon_{\text{BB}}(W) := (\pi_1\pi_2)^{\frac{1}{2}} \parent{\frac{\det{\frac{W^t\Sigma_1 W + W^t\Sigma_2 W}{2}}}{\sqrt{\det{W^t\Sigma_1 W} \det{W^t\Sigma_2 W}}}}^{-\frac{1}{2}} \, , 
\end{equation}
is $W_{\text{BB}}^* = [\phi_1, ..., \phi_q]$. See \cite{kadota1967best, fukunaga1972instruction} for more details. With this projection, the Bhattacharyya-overlap is:
\begin{equation} \label{eq:bhattacharyya_bound_optimal_embbeding}
    \epsilon_{\text{BB}}(W_{\text{BB}}^*) = (\pi_1\pi_2)^{\frac{1}{2}} \parent{\prod_{j=1}^q \frac{\lambda_j^{\frac{1}{2}} + \lambda_j^{-\frac{1}{2}}}{2}  }^{-\frac{1}{2}} \, .
\end{equation}

\subsubsection{Embedding dimension trade-off} \label{sect:q_tradeoff}

\noindent
{\bf Context.}
Consider a family of nested orthogonal projections $W_q := [w_1, ..., w_q] \in \R^{p \times q}$ for $q \in \llbrack{1, p}$. Let $f(\cdot; \theta) := \sum_k g(\cdot; \theta_k) : \R^q \longrightarrow \R_+$ be a parametric hierarchical pdf on $\R^q$, where the parameter $\theta:=(\theta_k)_{k=1}^K$ belongs to a certain parameter space $\Theta_q^K$. Call $z_{\theta}(\cdot):= \argmax{k} \brace{g(\cdot ; \theta_k)}$ the parametric model-based classifier naturally associated with $f$. In the finite sample case, we observe a dataset $X = (x_i^T)_{i=1}^n \in \R^{n \times p}$, drawn from a distribution which has {\it a priori} no connection to $f$. Still, we can learn a value for the parameter $\theta$ by solving a maximum likelihood optimisation problem of the form:
\begin{equation*}
    \widehat{\theta}(X, q) := \argmin{\theta \in \Theta_q^K} \brace{- \sum_i \ln(f(W_q^t x_i; \theta))} \, .
\end{equation*}
This problem can be supervised or unsupervised depending on whether the dataset $X$ is labelled or not. Regardless, the expected value, variance, in-sample goodness of fit and expected out-of-sample goodness of fit of the estimator $\widehat{\theta}(X, q)$ will depend on the sample size $n$ and the embedding dimension $q$. In turn, the classification error of the associated classifier $z_{\widehat{\theta}(X, q)}$ will also be affected by these  factors. Although increasing $n$ should always improve expected performance, the effect of $q$ is less clear. 

\medskip
\noindent
{\bf Optimal embedding dimension $q$.}
Consider the optimal embedding dimension with regards to the 0-1 loss:
\begin{equation*}
    q^*(X) := \argmin{1 \leq q \leq p} \brace{\Rcal\parent{z_{\widehat{\theta}(X, q)}}} \, .
\end{equation*}
This value of $q^*$ is not immediately obvious and indeed the choice of the embedding dimension $q$ 
not always discussed in detail. Yet, whether the learning procedure in $\R^q$ is supervised or not, the dimension $q$ plays a crucial role. Obviously, reducing $q$ reduces 
the dimension of the (embedded data) problem (and indeed this is the purpose of the projection procedure). Lower $q$ has a regularizing effect, e.g. better conditioned covariances \citep[see e.g.][]{rao1963discrimination, durrant2015random}. This added regularity can result in a better generalisation error. On the other hand, information is in general lost as we reduce $q$, potentially resulting in a trade-off.

In the remainder of this section, we argue that the behaviour of $q^*(X)$ is not trivial, as there is a trade-off between information-retention and regularisation to balance when setting $q$.

\medskip
\noindent
{\bf Loss of information when $q$ decreases.}
We have already seen that the optimal embedding size wrt Bayes risk $\epsilon(W)$, is always $q {=} p$, with $W {=} I_p$ (or any other orthogonal matrix in $\R^{p\times p}$). This is because the ambient space Bayes classifier $z^*(x) \! = \! \argmax{k} \brack{\P(z = k \vert  x)}$ has a lower risk than any other classifier, including any embedding Bayes classifier $z^*_W(x) \! = \! \argmax{k} \brack{\P(z = k \vert  W^t x)}$. To go even further, take our nested orthogonal projections $W_q = [w_1, ..., w_q]$ for $q \in \llbrack{1, p}$, and define the sequence of embedding Bayes classifiers:
\begin{equation} \label{eq:bayes_classifier_embedding_nested}
    z^*_q(x) := \argmax{k} \brack{\P(z = k \vert  W_q^t x)} \, .
\end{equation}
Then, by the same reasoning, we have a non-decreasing Bayes risk as the dimension decreases:
\begin{equation*}
    \Rcal(z^*_p) \leq \Rcal(z^*_{p-1}) \leq ... \leq \Rcal(z^*_{q}) \leq ... 
    \leq \Rcal(z^*_{1}) \, , 
\end{equation*}
where $z^*_p \! = \! z^*$. Working within the $q-$dimensional space defined by $W_q$ to build a classifier $\hat{z}_q : \R^q \rightarrow \llbrack{1, K}$, then it is always true that $\P(z^*_q(x) \neq z) \leq \P(\hat{z}_q(W_q^t x) \neq z)$, since $z^*_q$ implements the Bayes decision rule in $\mathrm{span}(W_q)$. Hence:
\begin{equation*}
    \Rcal(z^*_p) \leq \Rcal(z^*_{p-1}) \leq ... \leq \Rcal(z^*_{q}) \leq \Rcal(\hat{z}_q) \, . 
\end{equation*}
In short: a lower embedding dimension means worsening of the best potential performance. Hence, decreasing $q$ could lower predictive power.

\medskip
\noindent
{\bf Loss of stability when $q$ increases.}
On the other hand, higher $q$ also carries with it costs.  Even if we disregard added computational costs, statistical issues can arise when $q$ increases. Indeed, a criterion like the Bayes risk favours the highest dimensions because the Bayes estimator assumes exact knowledge of the model parameters. However in real applications, the data is finite, and the parameters have to be estimated. Consider for instance the centred Gaussian case: $\N(0_p, \Sigma_k)$. The Bayes decision rule \eqname\eqref{eq:bayes_classifier} uses the true value of the within-class pdf $p_k(x):=\P(x\vert z = k)$, which verifies:
\begin{equation*}
    - 2 \ln p_k(x) = x^t \Sigma_k^{-1} x + \ln \det{\Sigma_k}  + p\ln (2\pi)   \, .
\end{equation*}
In the case of the embedded Bayes classifier \eqref{eq:bayes_classifier_embedding_nested}, this becomes:
\begin{equation*}
    - 2 \ln p_k(W_q^t x) \equiv (W_q^t x)^t (W_q^t\Sigma_k W_q)^{-1} (W_q^t x) + \ln \det{W_q^t\Sigma_k W_q}  \, .
\end{equation*}
Where we omitted the additive constant $q\ln (2\pi)$ which is superfluous in the classification decision. In the subspace defined by the projection matrix $W_q$, $\Sigma_{k, q} := W_q^t \Sigma_k W_q$ is the covariance matrix of the embedded data  $x_q := W_q^t x$. With oracle knowledge of this matrix, one could compute the Bayes classifier \eqref{eq:bayes_classifier_embedding_nested}. However, in the finite sample case, $\Sigma_k$ is unavailable and must be replaced by an estimation. In the best case scenario, the available observations are labelled and can be split into $K$ datasets $X_k \in \R^{n_k \times p}$, one for each class $k$. With such information, we can compute the sample covariance matrices $S_k := {n_k}^{-1} X_k X_k^T$. 
Let $S_{k,q} :=  W_q^t S_k W_q \in S_q^{+}(\R)$. $S_{k,q}$ is full rank as long as $q \leq n_k$. In what follows, we will assume that all classes have the same sample size $n$ for the sake of simplicity. With $S_{k,q}$ at hand, we can define the finite sample version of the classifier \eqref{eq:bayes_classifier_embedding_nested} by replacing $\ln p_k(\cdot)$ with its sample analogue:
\begin{equation*} 
    - 2 \ln \hat{p}_k(x_q)\equiv x_q^t S_{k, q}^{-1} x_q + \ln \det{S_{k, q}}  \, .
\end{equation*}
Finally, the negative log-likelihood ratio between classes 1 and 2 can be written:
\begin{equation} \label{eq:empirical_classifier_embedding_nested}
\begin{split}
    \hat{r}_{1, 2} &:= - 2 \ln \frac{\hat{p}_1(x_q)}{\hat{p}_2(x_q)}\\
    &= x_q^t \parent{S_{1, q}^{-1} - S_{2, q}^{-1}} x_q + \ln \frac{\det{S_{1, q}}}{\det{S_{2, q}}}   \, .
\end{split}
\end{equation}
In a two-class problem, \eqname\eqref{eq:empirical_classifier_embedding_nested} is the criterion that decides whether $x$ is classified in class 1 or in class 2. However, the behaviour of this classifier becomes more unstable as $q$ increases towards any $n$. This is due to dependency of $\hat{r}_{1, 2}$ on both the inversion and the determinant of the two empirical covariance matrices $S_{1, q}$ and $S_{2, q}$, which are of rank $\mathrm{min}(q, n)$. As shown in \cite{edelman1991distribution}, with fixed sample size, the spectra of Wishart and inverse Wishart matrices gradually take more extreme values as the dimension grows. Hence, as $q$ increases, for a fixed data point $x$ and training sample size $n$, computing $\hat{r}_{1, 2}$ with different training sets will more easily lead to widely different values, potentially with different signs, leading to unstable labelling of the same data point $x$. In short, as $q$ grows, 
finite sample estimation effects can make the classifier unstable and potentially lead to higher risk.\\
\\
Overall, these results argue for treating $q$ as a tuning parameter that should be set in a data-adaptive manner. 
In particular, as can be seen in our experiments from Section \ref{sect:semi_theoretical} to \ref{sect:comparison_optimal}, following a simple heuristic like $q=K$ \citep{ding2004k}, with $K=2$ here, can lead to sub-optimal results (see \figurename~\ref{fig:finite_data_regret} for a particularly clear example).

\section{Does PCA projection preserve class separability?} \label{sect:PCA_theory}
The amount of signal retained by the projection depends not only on the embedding dimension $q$, but also on the projection methodology. For a fixed $q$, we have already described the Bhattacharyya-optimal projection in \sectionname~\ref{sect:optimal_subspace}. However, this projection can only be computed in a fully supervised setting, since it requires knowledge of the class-weights $\pi_k$ and the within-class covariances $\Sigma_k$. 
In the unsupervised setting, we are interested in projections that can be defined from only an unlabelled dataset $X \in \R^{n \times p}$. Random Projections (RP) are a very popular tool in this context \citep[see][and the many subsequent works]{dasgupta1999learning}. Yet, the RP are not trained and make no use of the information available in $X$. By contrast, Principal Component Analysis (PCA) extracts the dominant features of the full data $X$ and uses them to construct an ordered projection. Works such as \cite{fradkin2003experiments} and \cite{deegalla2006reducing} have already suggested that PCA projection performs better than RP in practice. 
We would like to understand the relative behaviour of PCA and RP in the second order, zero mean-difference case.


The $q-$dimensional PCA projection is defined by the equivalence class  of rectangular orthogonal matrices $W \! \in \! \R^{q \times p}$ that maximise the determinant of the embedded mixture-covariance: $\det{W^t (\Sigma_1 + \Sigma_2) W}$. Where we assumed balanced classes, and, to be as generous as possible, we consider the infinite sample size scenario, such that the true mixture-covariance matrix, $\Sigma_1 + \Sigma_2$, is available. The most natural solution is $W_q^{\text{PCA}}:=\brack{\phi_1, ..., \phi_q}$, made of the first $q$ eigenvectors $\phi_j$ of $\Sigma_1+\Sigma_2$.

On the other hand, as can be seen in \eqname\eqref{eq:bhattacharyya_bound_embbeding}, the Bhattacharyya-optimal projection maximises the same sum $\det{W^t (\Sigma_1 + \Sigma_2) W}$, while simultaneously keeping the product $\det{W^t\Sigma_1 W} \det{W^t\Sigma_2 W}$ minimal. This projection favours directions along which the two matrices have very different amplitudes (one large, one small). With no mean-signal, this is where the class-difference is most apparent.

Whether these two solutions are similar depends on the context. If the strongest eigenvectors $\brack{\phi_1, ..., \phi_q}$ of the sum $\Sigma_1 + \Sigma_2$ are directions alongside which only one matrix is strong while the other is weak (either $\phi_j^T \Sigma_1 \phi_j \gg \phi_j^T \Sigma_2 \phi_j$ or $\phi_j^T \Sigma_1 \phi_j \ll \phi_j^T \Sigma_2 \phi_j$), then the PCA choice is well suited to preserve class-separability. On the contrary, if both matrices have a similar amplitude in these directions ($\phi_j^T \Sigma_1 \phi_j \approx \phi_j^T \Sigma_2 \phi_j$), then the PCA choice is poorly suited to preserve class-separability. In the following, we express more formally this dichotomy and argue that there can be no meaningful bound on the PCA performance without assumptions on the matrices.

With two balanced latent subgroups, for any projection $W$, the Bayes risk trivially verifies $0\leq \epsilon(W) \leq 0.5$. Likewise, with $W_q^*$ the Bayes risk-optimal projection of size $q$, the regret $R(W):= \epsilon(W) - \epsilon(W_q^*)$ always verifies $0\leq R(W) \leq 0.5$. 
Tighter bounds could justify the decision to use or discard a specific projection. 
However, the following two counter-examples show that, without assumptions on $\Sigma_1$ and $\Sigma_2$, nontrivial bounds on the Bayes risk with PCA, $\epsilon(W_q^{\text{PCA}})$ or on the regret $R(W_q^{\text{PCA}})$ are not available.

\begin{example}[No lower bound exists]
    Let $0 < \delta < \alpha \in \R$. Consider a case where $\pi_1=\pi_2=0.5$ and the covariance matrices are:
    \begin{equation*}
        \Sigma_1 =\left[\begin{array}{c|c}
        \alpha I_q  & 0_{q \times (p-q)}\\ 
        \hline
        0_{(p-q)\times q} & \delta I_{p-q}
        \end{array}\right] \text{ and } \Sigma_2 = \delta I_p \, .
    \end{equation*}
    With such matrices, the feature selection rule of PCA coincides with the optimal decision rule. We have $W_q^{\text{PCA}} = W_q^* =: W_q$, with both selecting the first $q$ dimensions. Then $W_q^t \Sigma_1 W_q= \alpha I_q$ and $W_q^t  \Sigma_2 W_q= \delta I_q$, and the Bhattacharyya bound on the Bayes risk is:
    \begin{equation} \label{eq:bound_example}
        \epsilon(W_q) \leq \epsilon_{\text{BB}}(W_q) = \frac{1}{2} \parent{\prod_{j=1}^q \frac{(\frac{\alpha}{\delta})^{\frac{1}{2}} + (\frac{\delta}{\alpha})^{\frac{1}{2}}}{2}  }^{-\frac{1}{2}} \!\!\!\!\!\!\!\! \underset{\delta \longrightarrow 0}{\longrightarrow} 0 \, . 
    \end{equation}
    In this example, not only is the PCA regret always null, $R(W_q^{\text{PCA}}) = \epsilon(W_q^{\text{PCA}}) - \epsilon(W_q^*) = \epsilon(W_q) - \epsilon(W_q)=0$, but the PCA Bayes risk also converges towards 0 when $\delta \longrightarrow 0$. 
\end{example}

\begin{example}[No upper bound exists]
    We assume that the desired embedding space size verifies: $q \leq \frac{p}{2}$. This is very reasonable since, in most applications, any useful reduction of the dimension verifies: $q \ll p$. With $0 < \delta < \alpha \in \R$. Consider a case where $\pi_1=\pi_2=0.5$ and the covariance matrices are:
    \begin{equation*}
        \Sigma_1 =\left[\begin{array}{c|c}
        \alpha I_q  & 0_{q \times (p-q)}\\ 
        \hline
        0_{(p-q)\times q} & \delta I_{p-q}
        \end{array}\right] \text{ and } \Sigma_2 = \alpha I_p \, .
    \end{equation*}
    With such matrices, there is no overlap between the selection by PCA and the optimal selection. The PCA projection is made, as in the previous example, of the first $q$ directions, such that $W^{\text{PCA}\, t}_q \Sigma_1 W^{\text{PCA}}_q=W^{\text{PCA} \, t}_q  \Sigma_2 W^{\text{PCA}}_q= \alpha I_q$. Whereas the optimal projection selects $q$ components among the remaining $p-q\geq q$, such that $W^{*\, t}_q \Sigma_1 W^*_q= \delta I_q$ and $W^{*\, t}_q  \Sigma_2 W^*_q= \alpha I_q$. Since the two distributions are identical in the PCA embedding space, then $\epsilon(W^{\text{PCA}}_q) = 0.5$, the worst possible value for the Bayes risk. Whereas, with the optimal projection, the Bayes risk $\epsilon(W_q^*)$ verifies the bound \eqref{eq:bound_example}.
    
    In this example not only does the PCA Bayes risk always have the worst value possible of 0.5, but the regret of using PCA, $R(W_q^{\text{PCA}}) = \epsilon(W_q^{\text{PCA}}) - \epsilon(W_q^*)$ also converges towards the worst value of 0.5 when $\delta \longrightarrow 0$.
\end{example}

From these two examples we see that without assumption on $\Sigma_1$ and $\Sigma_2$, it is not possible to have finer bounds than 
$0 \leq \epsilon(W_q^{\text{PCA}}) \leq 0.5$ and $0 \leq R(W_q^{\text{PCA}}) \leq 0.5$.
The same is true of RP, since, in any circumstance, it can always by pure chance choose the exact same projections directions as PCA. This theoretical standoff motivates the use of quasi-exhaustive experiments to study the behaviour of PCA in this context.

\section{Quasi-exhaustive empirical analysis of the Bhattacharyya-overlap} \label{sect:semi_theoretical}
The unavailability of fully general results on the classification risk under PCA projection and RP motivates a more empirical approach. To begin with, in this section, we compare how successful PCA projection and RP are at reducing the Bhattacharyya-overlap \eqref{eq:bhattacharyya_bound_embbeding} between the two distributions. We assume Gaussian data $\N(0_p, \Sigma_k)$, and consider three different parametric families to draw the covariance matrices $(\Sigma_1, \Sigma_2)$ from. Thanks to the data-free closed form of the Bhattacharyya-overlap, we can efficiently valuate a very large number of pairs $(\Sigma_1, \Sigma_2)$. This allows us to explore in detail each matrix family.

\subsection{Strategy}
Theoretical results strive to cover analytically all possible cases. In this Section, we take an empirical approach with a 
similar, but less complete, ambition, namely to evaluate a large number of cases via the Bhattacharyya overlap. Assume that, for any $p\in \bN^*$, we have a parametric family $\F^{(p)}$ of distributions $\D^{(p)}_{\theta}$ over the space $S_p^{++}(\R)$ of covariance matrices of size $p$: $\F^{(p)}:=\brace{\D^{(p)}_{\theta}}_{\theta \in \Theta}$. Where $\theta$ is the parameter and $\Theta$ the parameter space. Let $n_p, n_{\theta}, n_{\text{simu}}, n_{\text{proj}}, n_q \in \bN^*$. We start by picking a value for $p$ among a given set of large integers $\{p_1, ..., p_{n_p}\}$. Then, we consider a finite grid $\widehat{\Theta}_{n_{\theta}} := \brace{\theta_1, ..., \theta_{n_{\theta}}} \subset \Theta$ over a bounded but large region of $\Theta$. For any combination $(\theta_i, \theta_j) \in \widehat{\Theta}_{n_{\theta}}^2$, we can independently generate any number $n_{\text{simu}}\in \bN^*$ of pairs of matrices $\Sigma_1 \sim \D^{(p)}_{\theta_i}$ and $\Sigma_2 \sim \D^{(p)}_{\theta_j}$. For every pair $(\Sigma_1, \Sigma_2)$, we then evaluate several different projection strategies (such as PCA projection, RP...) picked from a given set of $n_{\text{proj}}$ projections types. We also explore several values $\{q_1, ..., q_{n_q}\}$ of embedding space sizes. For any given projection type ``proj" and embedding size $q\leq p$, we compute projection $W_q^{\text{proj}}$ associated with the Mixture of Gaussian $\frac{1}{2} \N(0_p, \Sigma_1) + \frac{1}{2} \N(0_p, \Sigma_2)$. Finally, from $\Sigma_1, \Sigma_2$ and $W_q^{\text{proj}}$, we compute the Bhattacharyya-overlap $\epsilon_{\text{BB}}(W_q^{\text{proj}})$ in the projected subspace as in \eqname\eqref{eq:bhattacharyya_bound_embbeding}. In total, we generate $n_p \times n_{\theta}^2 \times n_{\text{simu}}$ independent pairs of matrices $(\Sigma_1, \Sigma_2)$, for a total of $n_p \times n_{\theta}^2 \times n_{\text{simu}} \times n_{\text{proj}} \times n_q$ different pairs of embedded matrices $(W_q^{\text{proj} \, t} \Sigma_1 W_q^{\text{proj}}, W_q^{\text{proj} \, t} \Sigma_2 W_q^{\text{proj}})$ to evaluate with the Bhattacharyya-overlap.

The ambition of this procedure is to be more than a simple ``localised" experiment. This is an almost exhaustive enumeration of the possible scenarios within a given family, which allows us to draw fairly general conclusions. The only barrier is the large number of experiments required. For reasonable grids, the number $n_p {\times} n_{\theta}^2 {\times} n_{\text{simu}} {\times} n_{\text{proj}} {\times} n_q$ can easily reach the tens or even hundreds of thousands. This is where the Bhattacharyya-overlap is useful. Indeed, thanks to its analytical nature, it be computed directly from $\Sigma_1, \Sigma_2$ and the projection $W$ (we do so below for large numbers of cases using only sequential programming and regular hardware)
and is hence computationally efficient relative to Monte Carlo estimation of the Bayes risk. 

In the three subsequent sections, we describe experiments with three different families of covariance matrices. A summary of the grid of settings explored in the experiments can be found in Table \ref{tab:experiment_settings}. All the experiment-specific parameters are properly introduced in the corresponding section.

\begin{table}[tbhp]
\caption{Recapitulation of the experimental settings. For each combinations of these settings, we independently generate 100 matrix pairs Exp. \ref{sect:exp_IW} and \ref{sect:exp_latent_IW}, and 20 matrix pairs for Exp. \ref{sect:exp_RNA}.  } \label{tab:experiment_settings}
\begin{tabular}{l|c|c|c}
\hline
parameters & Exp. \ref{sect:exp_IW}: IW & Exp. \ref{sect:exp_latent_IW}: Latent low-dim. IW & Exp. \ref{sect:exp_RNA}: Real data \\ \hline
$p$ & 20, 50, 100, 200 & 20, 50, 100, 200 & 100, 500, 1000 \\
$q$ & 1, 2, 5, 10, 50  & 1, 2, 5, 10, 50  & 2, 5, 10, 20 \\ \hline
$df_1/p$ & 1, 1.5, 2, 3, 4, 5, 10 & \cellcolor[HTML]{A5A5A5}{\color[HTML]{FFFFFF} } & \cellcolor[HTML]{A5A5A5}{\color[HTML]{FFFFFF} } \\
$df_2/p$ & 1, 1.5, 2, 3, 4, 5, 10 & \cellcolor[HTML]{A5A5A5}{\color[HTML]{FFFFFF} } & \cellcolor[HTML]{A5A5A5}{\color[HTML]{FFFFFF} } \\ \hline
mixing matrices $Q$                    & \cellcolor[HTML]{A5A5A5}{\color[HTML]{FFFFFF} } & dense, sparse & \cellcolor[HTML]{A5A5A5}{\color[HTML]{FFFFFF} } \\
$(\indic_{Q_1=Q_2}, \indic_{\Theta_1=\Theta_2})$ & \cellcolor[HTML]{A5A5A5}{\color[HTML]{FFFFFF} } & (0, 0), (1, 0), (0, 1) & \cellcolor[HTML]{A5A5A5}{\color[HTML]{FFFFFF} }\\ \hline
$\gamma$ & \cellcolor[HTML]{A5A5A5}{\color[HTML]{FFFFFF} } & \cellcolor[HTML]{A5A5A5}{\color[HTML]{FFFFFF} } & $0, \frac{1}{8}, \frac{2}{8}, \frac{3}{8}, \frac{4}{8}, \frac{5}{8}, \frac{6}{8}, \frac{7}{8}, 1$ \\ \hline
nb combinations & 882 & 108 & 108 \\ \hline
nb matrix pairs & 88200 & 10800 & 2160 \\ \hline
\end{tabular}
\end{table}

\subsection{Family 1: inverse Wishart}\label{sect:exp_IW}
First, we consider an inverse Wishart (IW) distribution family. Let $p \in \bN^*$ be fixed, we note $\Sigma {\sim} \D^{(p)}_{df}$ when $\Sigma {:=} df \times W$, where $W {\sim} \mathcal{W}^{-1}(I_p, df)$ is an inverse Wishart matrix with scale matrix $I_p$ and $df\geq p$ degrees of freedom. The multiplication by $df$ keeps the scale of the matrices consistent even when $p$ is large. For our simulation, we generate our pairs $(\Sigma_1, \Sigma_2)$ of independent matrices as $\Sigma_1 {\sim} \D^{(p)}_{df_1}$ and $\Sigma_2 {\sim} \D^{(p)}_{df_2}$ with $df_1$ and $df_2$ two integers. The grid of hyper-parameters we consider is: $(p, q) \in [20, 50, 100, 200]\times [1, 2, 5, 10, 50]$, and for a given value of $p$: $(df_1, df_2) \in [p, 1.5p, 2p, 3p, 4p, 5p, 10p]^2$. We consider all the possible combinations of parameters on this grid, excluding only the combinations where $q>p$. For each hyper-parameter combination, we simulate $n_{\text{simu}} = 100$ different pairs of covariance matrices. For each pair, we consider three projections: PCA projection, RP and sparse-RP. With PCA projection, $W$ is made of the $q$ largest eigenvectors of $\Sigma_1+\Sigma_2$. Each RP simply generates a random matrix $W \in \R^{p \times q}$. For the regular RP, the entries are iid Gaussian: $W_{ij} {\sim} \N(0, 1)$. For sparse-RP, we implement the \textit{very sparse} form of \cite{li2006very}: the entries can take one of three value $\{-p^{\frac{1}{4}}, 0, p^{\frac{1}{4}}\}$, with respective probabilities $\{\frac{1}{2 \sqrt{p}}, 1-\frac{1}{\sqrt{p}}, \frac{1}{2 \sqrt{p}}\}$. The generated $W$ is redrawn if not of rank $q$.

\begin{figure*}[tbhp]
    \centering
    \includegraphics[width=\linewidth]{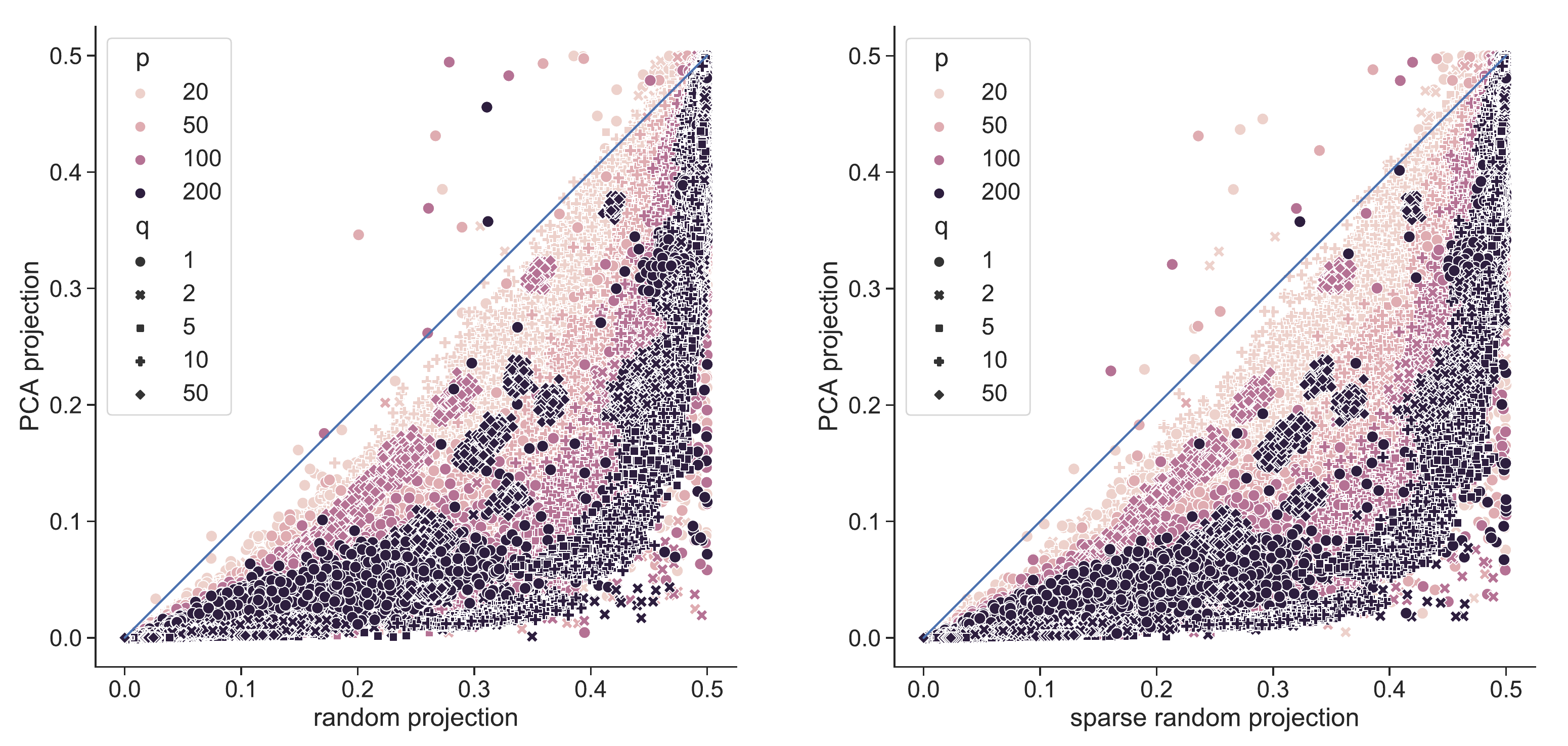}
    \caption{\textbf{Inverse Wishart generation scheme}. Bhattacharyya-overlap within the embedding subspace. PCA ($y$-axis) vs random projections ($x$-axes). Dense RP on the left, sparse-RP on the right. The overlap is better with PCA in $98.7\%$ of the cases. A total of 88200 pairs $(\Sigma_1, \Sigma_2)$ of ambient space matrices were generated for the experiment.}
    \label{fig:scatterplot_IW}
\end{figure*}

In total, 88200 different pairs $(\Sigma_1, \Sigma_2)$ of ambient space matrices are generated, and 264600 pairs of embedded matrices are evaluated. Each of the 88200 individual simulations is represented as one dot on \figurename~\ref{fig:scatterplot_IW}. The two $(x, y)$ coordinates of each dot are the values of the Bhattacharyya-overlap within two different embedding spaces. On both sub-figures, the $y$-coordinate corresponds to the embedding space defined by the PCA projection. The $x$-coordinate corresponds to the random projection on the left sub-figure and to the sparse random projection on the right. In $98.7\%$ of the cases, the two distributions are better Bhattacharyya-separated within the PCA subspace than the random subspaces. This striking figure indicates a strong domination of PCA embeddings over random embeddings in the IW case.

\subsection{Family 2: latent low-dimension inverse Wishart}\label{sect:exp_latent_IW}
Then, we consider a matrix generation scheme where the covariance matrices have a latent low-dimension structure. We take $r:=\text{max}(2, p/25)$ as the low latent dimension, and for each class $k=1,2$, we define a latent covariance matrix $\Theta_k {\sim} \mathcal{W}^{-1}(I_r, r+1)$, a mixing matrix $Q_k \in \R^{r \times p}$, a noise matrix $M_k {\sim}  \mathcal{W}^{-1}(I_p, 2p) $ and define: $$\Sigma_k := (r+1) Q_k^T \Theta_k Q_k  + 0.02 p M_k.$$ Note that, although a latent linear model, these data arise from mixture of subspaces and not a PCA model, hence may data  not be particularly well suited to PCA.
For the simulation, we consider as before several combinations of initial and target dimensions $(p, q) \in [20, 50, 100, 200] \times [1, 2, 5, 10, 50]$. We only consider $(p, q)$ combinations where $q<p$. For each of these combinations, we consider three different configurations of the generation scheme: one where both the mixing matrices and the latent covariance matrices are different ($Q_1{\neq} Q_2$ and $\Theta_1 {\neq} \Theta_2$), one where the mixing matrices are the same and the latent covariances ifferent ($Q_1 {=} Q_2$ and $\Theta_1 {\neq} \Theta_2$), and one where the mixing matrices are different and the latent covariances  the same ($Q_1 {\neq} Q_2$ and $\Theta_1 {=} \Theta_2$). We never consider at the same time $Q_1 {=} Q_2$ and $\Theta_1 {=} \Theta_2$, to avoid the case where $\Sigma_1$ and $\Sigma_2$ are almost identical. For each of these three cases, we consider one variant where the mixing matrices $Q_k$ are sparse, and one where there are dense. This results in a combination of six different generation scheme scenarios. As previously, we generate $n_{\text{simu}}{=}100$ pairs of covariance matrices for each parameter configuration, and evaluate three projection types: PCA, RP and sparse-RP.

\begin{figure*}[tbhp]
    \centering
    \includegraphics[width=\linewidth]{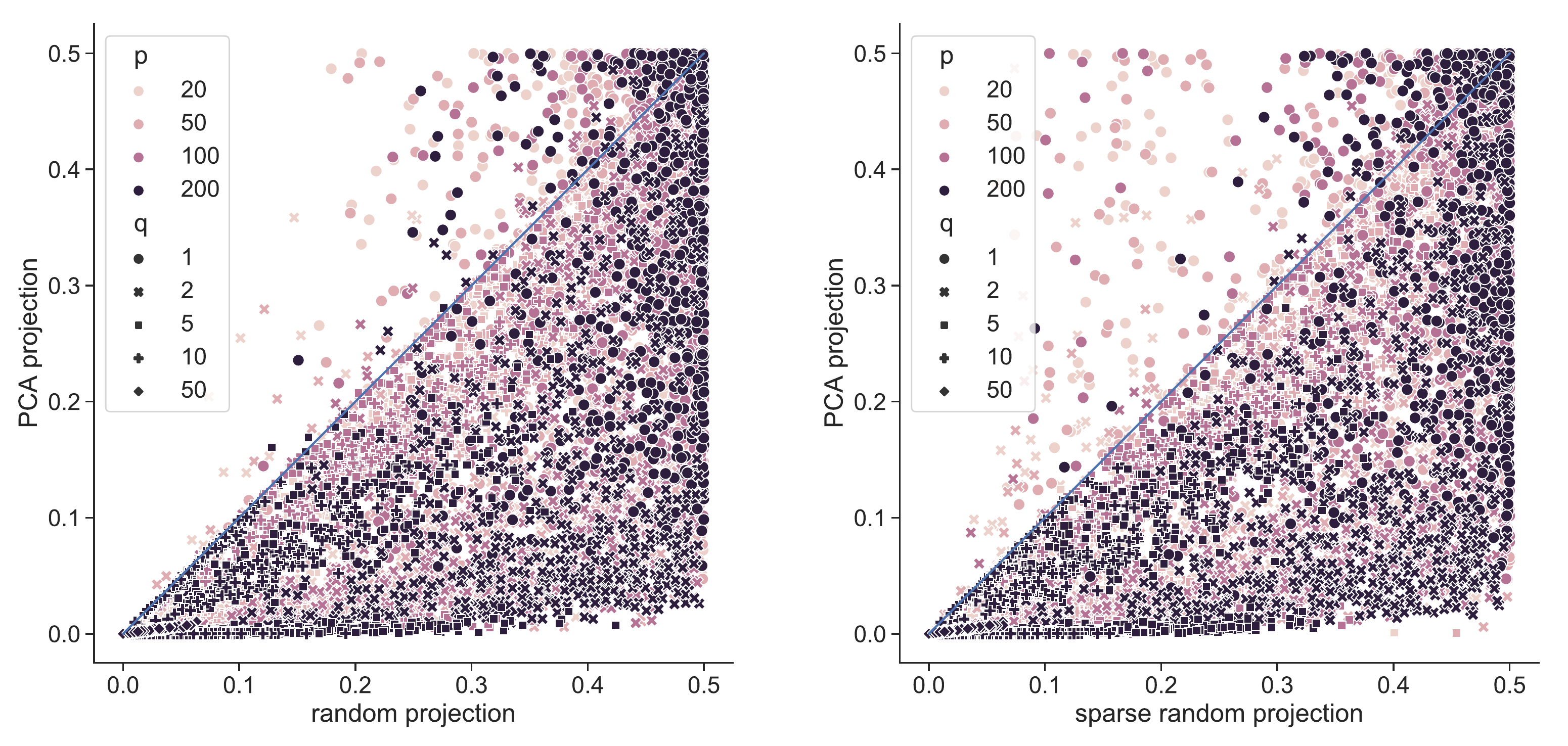}
    \caption{\textbf{Latent low-dimension inverse Wishart generation scheme}. Bhattacharyya-overlap within the embedding subspace. PCA ($y$-axis) vs random projections ($x$-axes). Dense RP on the left, sparse-RP on the right. The overlap with PCA is better than with RP in $94.4\%$ of the cases, and better than with sparse-RP in $93.5\%$ of the cases. A total of 10800 pairs $(\Sigma_1, \Sigma_2)$ of ambient space matrices were generated for the experiment.}
    \label{fig:scatterplot_latent_space}
\end{figure*}

In total, 10800 different pairs $(\Sigma_1, \Sigma_2)$ of ambient space matrices are generated and 32400 pairs of embedded matrices are evaluated. We depict the results on \figurename~\ref{fig:scatterplot_latent_space} with the same conventions as \figurename~\ref{fig:scatterplot_IW}. As previously, the class-separability in the PCA embedding space is much better. PCA is better than RP in $94.4\%$ of the cases, and better than sparse-RP in $93.5\%$.

\subsection{Family 3: empirical covariance of real RNA-seq data}\label{sect:exp_RNA}
Finally, we consider a family of matrices where each member is generated as the empirical covariance of a real dataset. From a centred dataset $X \in \R^{n \times p}$ with $n \in \bN^*$ observations, we can define a symmetric non-negative matrix by taking the empirical covariance $\Sigma:={n}^{-1} X^T X$. By considering highly non-Gaussian data, and datasets where the number of observations $n$ is either comparable to or smaller than the dimension $p$, we hope to generate covariance matrices which are very different from inverse Wishart or other matrices generated from theoretical probability distributions. Additionally, note that in the previous two examples, the two matrices $(\Sigma_1, \Sigma_2)$ in a pair were always generated independently from one another. This could lead to an over-representation of pairs of very dissimilar matrices (although, in the IW case, when both $df_1$ and $df_2$ are large, the two matrices are almost identical $=I_p$). Since we want to cover as many different scenarios as possible, we introduce in this example a transformation, detailed below, that enforces a certain form of overlap between some of the dimensions of the two matrices. Finally, we recall that in section 
we study the theoretical separability between two matrices $\Sigma_1$ and $\Sigma_2$ under the Gaussian assumption. Hence, the data $X$ is only used here as a means to generate a new type of matrices $\Sigma$ and is discarded afterwards. In particular, we do not use $X$ to compute any notion of classification error here. See Section \ref{sect:classifier_error} for a full analysis of classifiers trained with finite real data an evaluated on Out of Sample (OoS) data.

We use single-cell RNA-sequencing dataset used in \cite{taschler2019model}, with a total of approximately $18000$ measured genes and where there are two physically separate cell populations taken from two different areas of the brain (dorsal root ganglia and hippocampus). There are $n_1{=}290$ cells from group 1 and $n_2{=}390$ cells from group 2. For our simulations, we can work with any dimension $p<18000$ by sub-sampling without replacement $p$ columns among the total 18000 available. We call $X_1 \in \R^{n_1 \times p}$ and $X_2 \in \R^{n_2 \times p}$ the sub-sampled datasets corresponding to cell groups 1 and 2.

\begin{figure}[tbhp]
    \centering
    \includegraphics[width=0.6\linewidth]{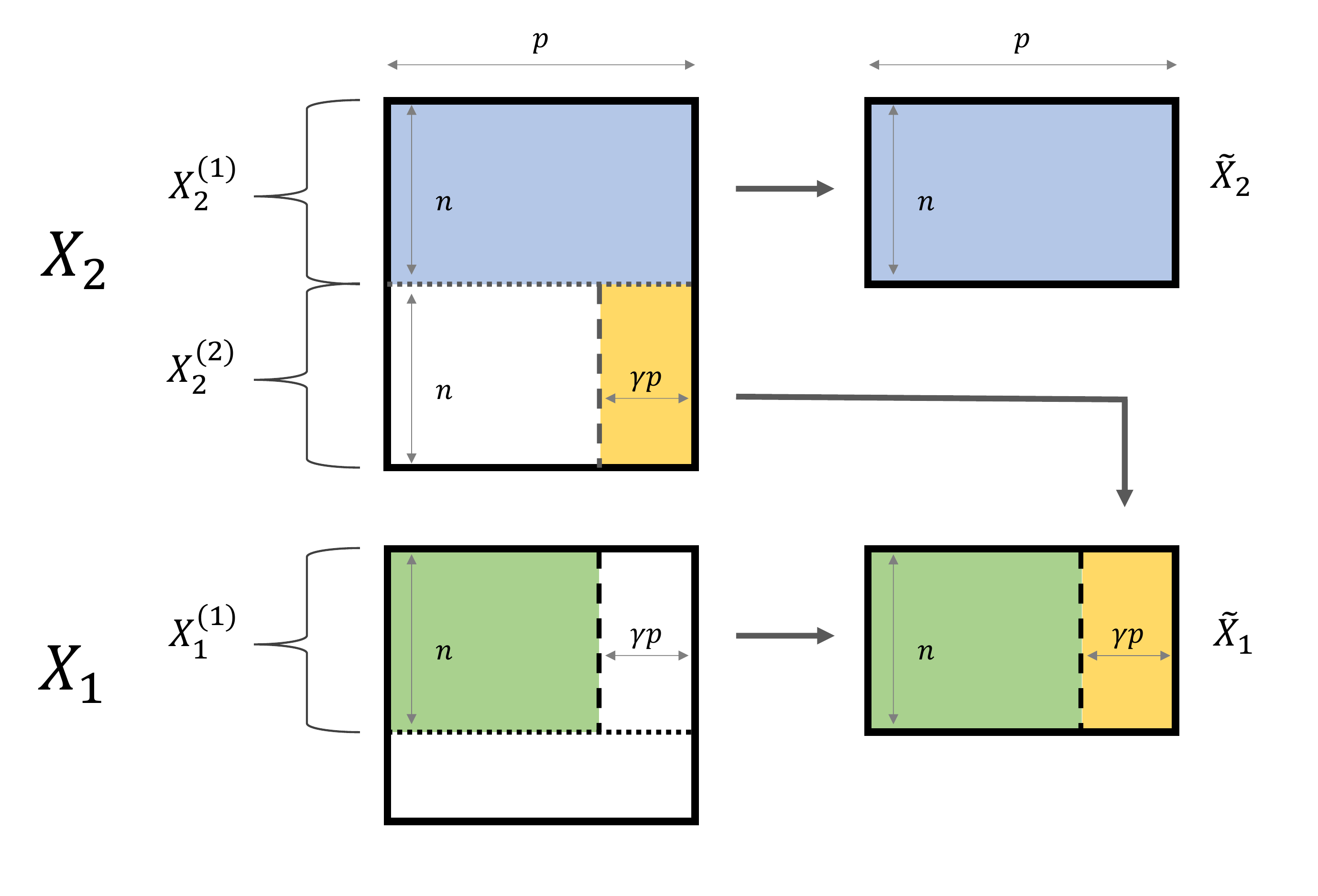}
    \caption{Illustration of the \textit{column overlap} procedure. The partition of the rows of $X_1$ and $X_2$ as well as the selection of the $\lfloor\gamma p\rfloor$ columns that overlap are all at random.}
    \label{fig:column_overlap}
\end{figure}

The two groups have real physiological differences, hence their corresponding distributions are expected to be very different, resulting in independent $\Sigma_1$ and $\Sigma_2$. To exert more control on their similarity, we propose a procedure that makes the distributions of a certain subset of the columns of $X_1$ and $X_2$ overlap. Consider a proportion of overlap $\gamma \in [0, 1]$ such that $\lfloor\gamma p\rfloor \in \bN$ is the target number of overlapping columns. For each simulation, the most numerous group, group 2, is randomly divided in two sub-groups $X_2^{(1)}$ and $X_2^{(2)}$ with $n{:=} 0.5\, n_2 {=}195$ samples each. Likewise, $n {=} 195$ rows are randomly sub-sampled from $X_1$ to form a dataset $X_1^{(1)}$. Then $\lfloor\gamma p\rfloor$ columns are randomly picked to be the overlapping ones. We replace the value of these columns in $X_1^{(1)}$ by the values of the corresponding columns of $X_2^{(2)}$. We call $\widetilde{X}_1$ the resulting dataset. In the end, the datasets used for the rest of the algorithm are $\widetilde{X}_1 \in \R^{n \times p}$ and $\widetilde{X}_2 := X_2^{(1)} \in \R^{n \times p}$. See \figurename~\ref{fig:column_overlap} for a visual illustration of the process. With this procedure, $\widetilde{X}_1$ and $\widetilde{X}_2$ remain independent, none of their columns are actually equal. Instead, the overlapping columns simply follow the same distribution. By increasing the overlap parameter $\gamma$, we can make the distributions and covariance matrices of $\widetilde{X}_1$ and $\widetilde{X}_2$ gradually more similar, from describing two different population when $\gamma = 0$ to two identical populations when $\gamma = 1$.

After the sub-sampling and overlap tuning steps are done, we compute the two empirical covariance matrices: $\Sigma_1{:=}{n_1}^{-1}\widetilde{X}_1^T \widetilde{X}_1$ and $\Sigma_2{:=}{n_2}^{-1}\widetilde{X}_2^T \widetilde{X}_2$. Then, as mentioned, we discard the real data points, and the generated matrices $(\Sigma_1, \Sigma_2)$ are used to define two centred Gaussian distributions: $\N(0_p, \Sigma_1)$ and $\N(0_p, \Sigma_2)$. As with the previous two matrix families, we use $(\Sigma_1, \Sigma_2)$ to compute the PCA projection and the embedded Bhattacharyya-overlap under the Gaussian assumption. Over our different simulations, we consider the grid of settings $(p, \gamma, q) = [100, 500, 1000]\times[0, 0.125, 0.25, 0.375, 0.5, 0.625, 0.75, 0.875, 1] \times [2, 5, 10, 20]$. We generate $n_{\text{simu}}{=}20$ pairs of covariance matrices for each parameter configuration, and evaluate the same three projection types as before.

\begin{figure*}[tbhp]
    \centering
    \includegraphics[width=\linewidth]{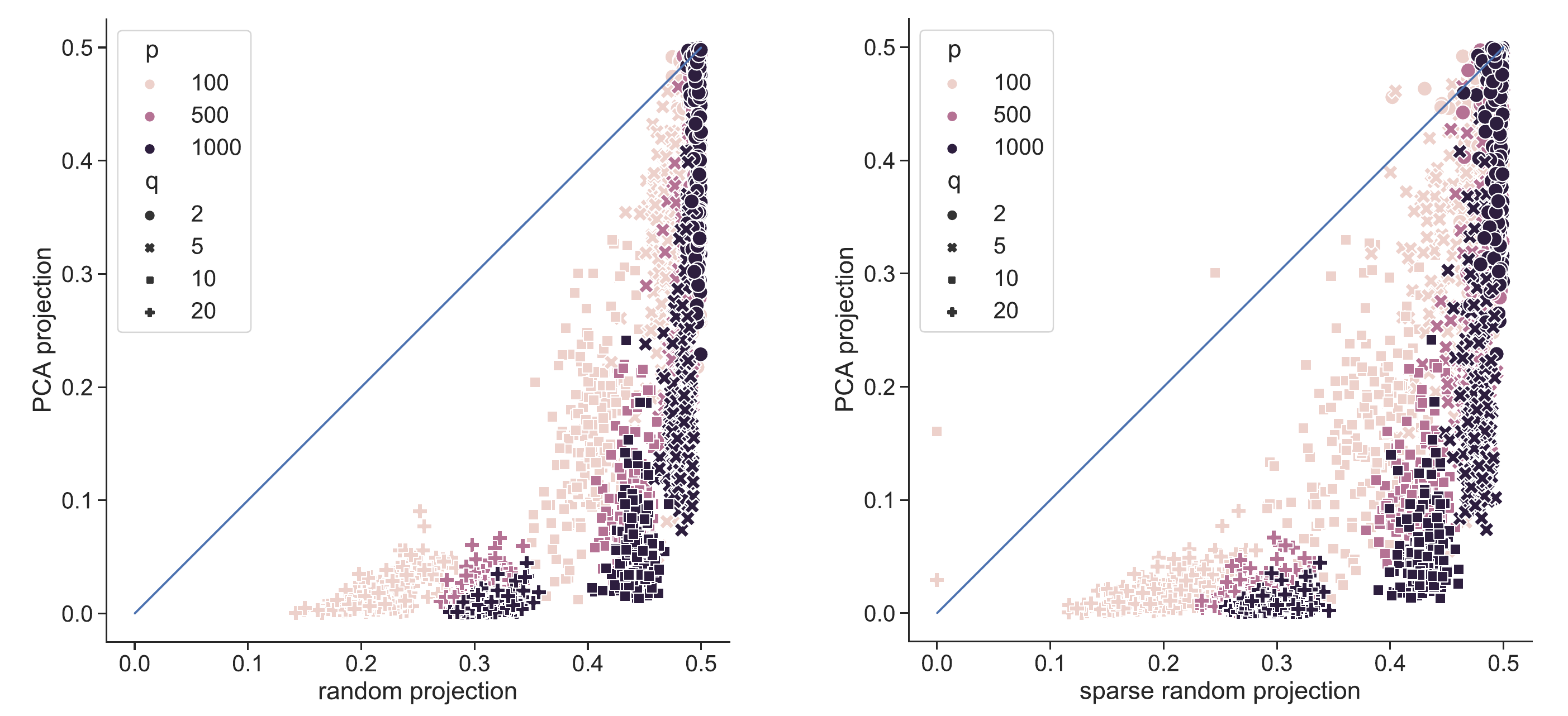}
    \caption{\textbf{Empirical covariance of real data generation scheme (here: single cell RNA sequencing data)}. Bhattacharyya-overlap within the embedding subspace. PCA ($y$-axis) vs random projections ($x$-axes). Dense RP on the left, sparse-RP on the right. The overlap with PCA is better than with RP in $98.8\%$ of the cases, and better than sparse-RP in $97.3\%$ of the cases. See Table \ref{tab:bhattacharyya_regret}. A total of 2160 pairs $(\Sigma_1, \Sigma_2)$ of ambient space matrices were generated for the experiment.}
    \label{fig:scatterplot_RNA_seq}
\end{figure*}

In total, 2160 different pairs $(\Sigma_1, \Sigma_2)$ of ambient space matrices are generated and 6480 pairs of embedded matrices are evaluated. We represent the results on \figurename~\ref{fig:scatterplot_RNA_seq} with the same conventions as before. For this family of matrices as well, we observe that the PCA projection is better suited to preserve the class separability in low dimension than the random projections. Overall, PCA is better than RP in $98.8\%$ of the cases, and better than sparse-RP in $97.3\%$. In Table \ref{tab:bhattacharyya_regret}, we propose a finer analysis over the of the column overlap $\gamma$. We display the average values of the Bhattacharyya-overlap with PCA projection, $\epsilon_{\text{BB}}(W^{\text{PCA}})$, and of the regret of both RP with regards to PCA, $r(W^{\text{RP}}) := \epsilon_{\text{BB}}(W^{\text{RP}})-\epsilon_{\text{BB}}(W^{\text{PCA}})$, as well as the average number of times that the regret is positive, $\indic_{r(W^{\text{RP}})>0}$. These are overall numbers, aggregated over all simulations and values of $(p, q)$. According to the Bhattacharyya-overlap, PCA projection broadly outperforms the RPs for all $\gamma$.

\begin{table*}[tbhp]
\caption{For each level of overlap $\gamma$ between the columns of $X_1$ and $X_2$, average values of the Bhattacharyya-overlap of PCA projection $\epsilon_{\text{BB}}(W^{\text{PCA}})$, of the regrets of both RP, and of the number of times these regrets are positive. According to the Bhattacharyya-overlap, the regrets of the RP are consequent for all values of $\gamma$, and choosing PCA is almost always the best option.} \label{tab:bhattacharyya_regret}
\begin{tabular}{l|c|cc|cc}
\hline
$\gamma$ & $\epsilon_{\text{BB}}(W^{\text{PCA}})$ & $r(W^{\text{RP}})$ & $\indic_{r(W^{\text{RP}})>0}$ & $r(W^{\text{sp.-RP}})$ & $\indic_{r(W^{\text{sp.-RP}})>0}$  \\ \hline
0              & 0.19  & 0.23  & 1.00  & 0.21  & 0.98   \\ 
0.125          & 0.19  & 0.23  & 1.00  & 0.22  & 0.98   \\ 
0.25           & 0.18  & 0.24  & 1.00  & 0.22  & 0.98   \\ 
0.375          & 0.17  & 0.25  & 0.99  & 0.23  & 0.99   \\ 
0.5            & 0.17  & 0.25  & 1.00  & 0.24  & 1.00   \\ 
0.625          & 0.17  & 0.24  & 1.00  & 0.23  & 0.99   \\ 
0.75           & 0.20  & 0.22  & 0.98  & 0.21  & 0.98   \\ 
0.875          & 0.23  & 0.19  & 0.97  & 0.18  & 0.95   \\ 
1              & 0.24  & 0.18  & 0.95  & 0.17  & 0.92   \\ \hline
\end{tabular}
\end{table*}

\section{Real data experiments with classification error}\label{sect:classifier_error}
In \sectionname~\ref{sect:semi_theoretical}, we showcased on a very wide range of matrix pairs, how PCA projection defines sub-spaces with better Bhattacharyya-overlap than RP. However, how does this translate in terms of actual 0-1 loss? Indeed, although the Bhattacharyya-overlap controls the classification risk as per Eq.~\eqref{eq:bhattacharyya_bound_general}, it still is only an upper bound on said risk. Moreover, the bound is derived under a Gaussian assumption. To address this question, in this section, we explore the classification risk under each projection in the finite sample, non-Gaussian case.

We compare the Out of Sample (OoS) classification error of classifiers trained on data embedded either within PCA projection sub-spaces or within RP sub-spaces. Let $X_{\text{train}} \in \R^{n_{\text{train}} \times p}$ be a labelled training set. The PCA projection $W^{\text{PCA}}$ is learned on the full data (without using the labels). 
For any projection $W {\in} \R^{p \times q}$, the projected train dataset is $X_{\text{train}} W \! \in \! \R^{n_{\text{train}} \times q}$. Using knowledge of the labels, we compute the empirical class weights $\widehat{\pi}_k$, embedding average $ W^t \widehat{\mu}_k \! \in \! \R^q$ and embedding covariance $W^t \widehat{\Sigma}_k W \! \in \! S_q^{++}(\R)$ from $X_{\text{train}}  W$ and  define an embedded likelihood ratio classifier $z$ as in \eqname\eqref{eq:bayes_classifier_embedding} with these parameters:
\begin{equation} \label{eq:trained_classifier_embedding}
    \widehat{z}_{W}(x) := \argmax{k}\brack{ \widehat{\pi}_k  \varphi(W^t x ; W^t \widehat{\mu}_k, W^t\widehat{\Sigma}_k W)} \, .
\end{equation}
With OoS validation data $(Z_{\text{val}}, X_{\text{val}}) \in \{1, 2\}^{n_{\text{val}}} \times \R^{n_{\text{val}} \times p}$, we can compute the OoS error (0-1 loss):
\begin{equation} \label{eq:empirical_risk}
    \frac{1}{n_{\text{val}}} \sum_{i = 1}^{n_{\text{val}}} \indic_{Z_{\text{val}, \, i} = \widehat{z}_{W}(X_{\text{val}, \, i})} \, .
\end{equation}
We compare the OoS error obtained with PCA projection, RP and sparse-RP on the RNA sequencing dataset. $70\%$ of the data is used in the training set, and $30\%$ in the validation set. To cover as much ground as possible, we also include the \textit{column overlap} procedure of real data experiments of Section \ref{sect:semi_theoretical}. We also use the same combination of parameters, $(p, \gamma, q) = [100, 500, 1000]\times[0, 0.125, 0.25, 0.375, 0.5, 0.625, 0.75, 0.875, 1] \times [2, 5, 10, 20]$, and the same number of simulations, $n_{\text{simu}} = 20$, for each configuration. We recall that with this data transformation, each class has $n_1=n_2=195$ data points, for a total sample size of $n = 390$. 

\begin{remark}
Note that a supervised classifier is used here, even though our ultimate goal is unsupervised learning. Indeed, the scope of this study is to evaluate and compare the projection methods. By following up the projection step with a supervised classifier, we consider the ``best case performances" that any clustering method could reach. 
We refer to \cite{taschler2019model} for a study that tackles the comparison between and choice amongst several projected clustering methods in a fully unsupervised setting.
\end{remark}

\begin{figure*}[tbhp]
    \centering
    \includegraphics[width=\linewidth]{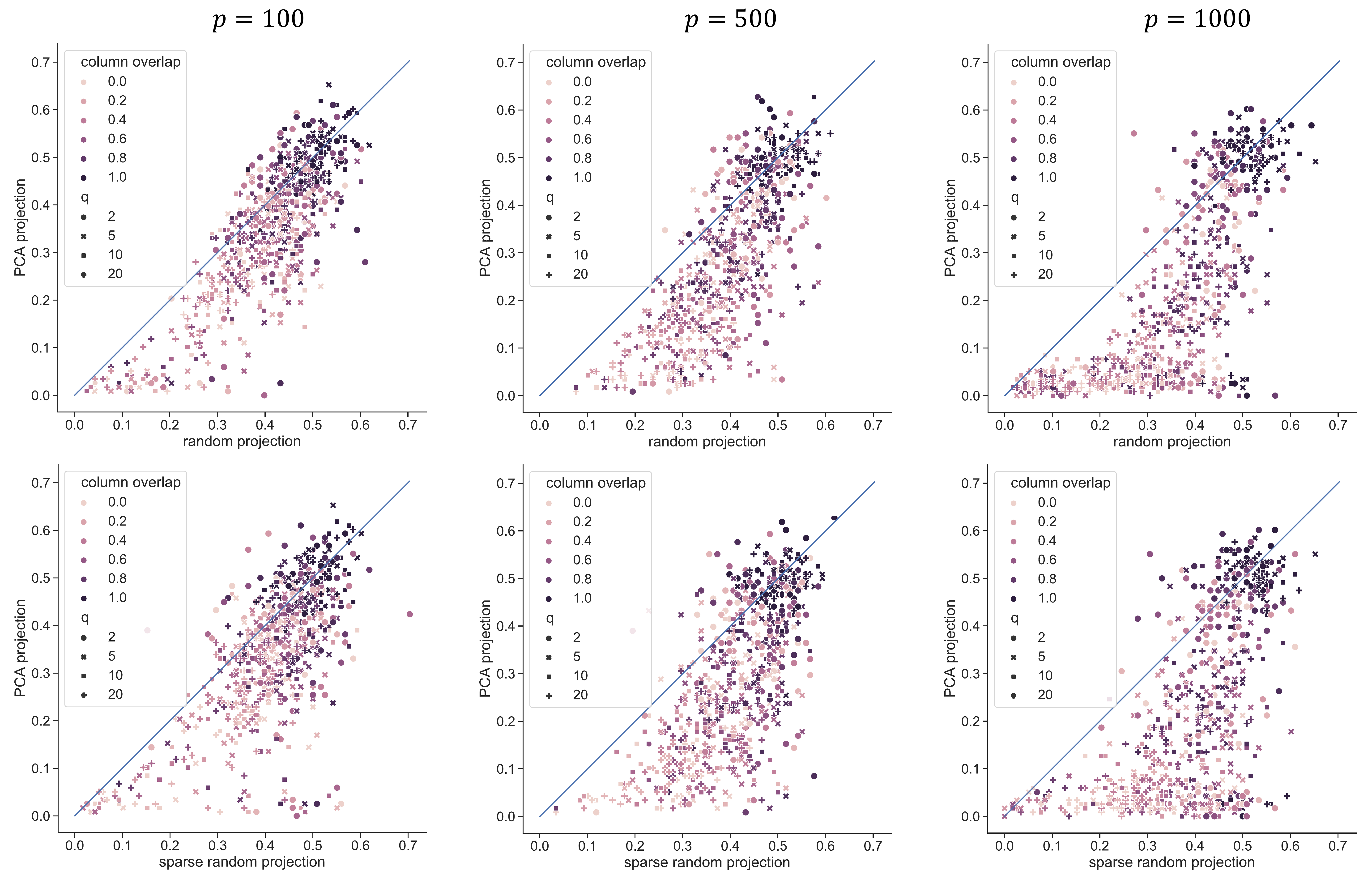}
    \caption{Out of Sample classification error of a $q-$dimensional model based classifier. PCA ($y$-axis) vs random projections ($x$-axes), stratified by ambient sizes $p$. The domination of PCA, although less strong than on the theoretical \figurename~\ref{fig:scatterplot_RNA_seq}, is still clear. We also observe that the comparative advantage of PCA is more pronounced for higher ambient dimensions $p$.}
    \label{fig:finite_data_classifier}
\end{figure*}

Each of the 6480 outcomes of this simulation are depicted on \figurename~\ref{fig:finite_data_classifier}. This time we make a sub-figure for each value of the ambient size $p$ and colour the data points according to the value of the column overlap $\gamma$. As expected, $\gamma$ has an impact on performance: when $\gamma \! \longrightarrow \! 1$, correct classification becomes gradually harder until it is impossible. Although the situation is less clear cut that in Section \ref{sect:semi_theoretical}, we can see that the OoS error is more often lower within the PCA embedding sub-space. To gain further insight, in Table \ref{tab:finite_data_regret}  we show the average OoS 0-1 loss $l(W^{\text{PCA}})$ with the PCA projection, the average regret $r(W^{\text{RP}}) \! := \! l(W^{\text{RP}}) {-} l(W^{\text{PCA}})$ of both RP, and the average number of times that these regrets are positive, $\indic_{r(W^{\text{RP}})>0}$. All these numbers are computed as a function of the overlap $\gamma$ between columns of $X_1$ and $X_2$, and averaged over all simulations and values of $(p, q)$. Table \ref{tab:finite_data_regret} is to be put in perspective with Table \ref{tab:bhattacharyya_regret}, its theoretical counterpart involving the Bhattacharyya-overlap. Overall, we observe that PCA projection is less overwhelmingly dominant in terms of 0-1 loss than it was in terms of Bhattacharyya-overlap. In particular, the 0-1 loss regret of choosing (any of the two) RP over PCA is lower in average than the corresponding Bhattacharyya-overlap regret. Still, the average signs of the regret on Table \ref{tab:finite_data_regret}, show that PCA projection retains a much higher chance of being the better choice overall. 
Moreover, the average value of these regrets are also non-negligible when compared to $l(W^{\text{PCA}})$, which further justifies favouring PCA projection.

\begin{table*}[tbhp]
\caption{For each level of overlap $\gamma$ between the columns of $X_1$ and $X_2$, average of the 0-1 loss $l(W)$ with the PCA, of the regrets of both RP and of the number of times these regrets are positive. These regrets are lower than their theoretical counterparts from Table \ref{tab:bhattacharyya_regret}, but still consequent when compared to PCA 0-1 loss, especially when $\gamma$ is small. Moreover, PCA is still largely preferable giving the signs of the regret.} \label{tab:finite_data_regret}
\begin{tabular}{l|c|cc|cc}
\hline
$\gamma$ & $l(W^{\text{PCA}})$ & $r(W^{\text{RP}})$ & $\indic_{r(W^{\text{RP}})>0}$ & $r(W^{\text{sp.-RP}})$ & $\indic_{r(W^{\text{sp.-RP}})>0}$  \\ \hline
0              & 0.22  & 0.13  & 0.91  & 0.15  & 0.90   \\ 
0.125          & 0.24  & 0.12  & 0.91  & 0.15  & 0.90   \\ 
0.25           & 0.23  & 0.12  & 0.89  & 0.15  & 0.91   \\ 
0.375          & 0.25  & 0.12  & 0.90  & 0.14  & 0.88   \\ 
0.5            & 0.24  & 0.13  & 0.91  & 0.16  & 0.91   \\ 
0.625          & 0.28  & 0.12  & 0.88  & 0.14  & 0.90   \\ 
0.75           & 0.31  & 0.10  & 0.82  & 0.12  & 0.89   \\ 
0.875          & 0.37  & 0.07  & 0.75  & 0.09  & 0.77   \\ 
1              & 0.49  & 0.02  & 0.58  & 0.02  & 0.61   \\ \hline
\end{tabular}
\end{table*}

\begin{figure*}[tbhp]
    \centering
    \subfloat{\includegraphics[width=\linewidth]{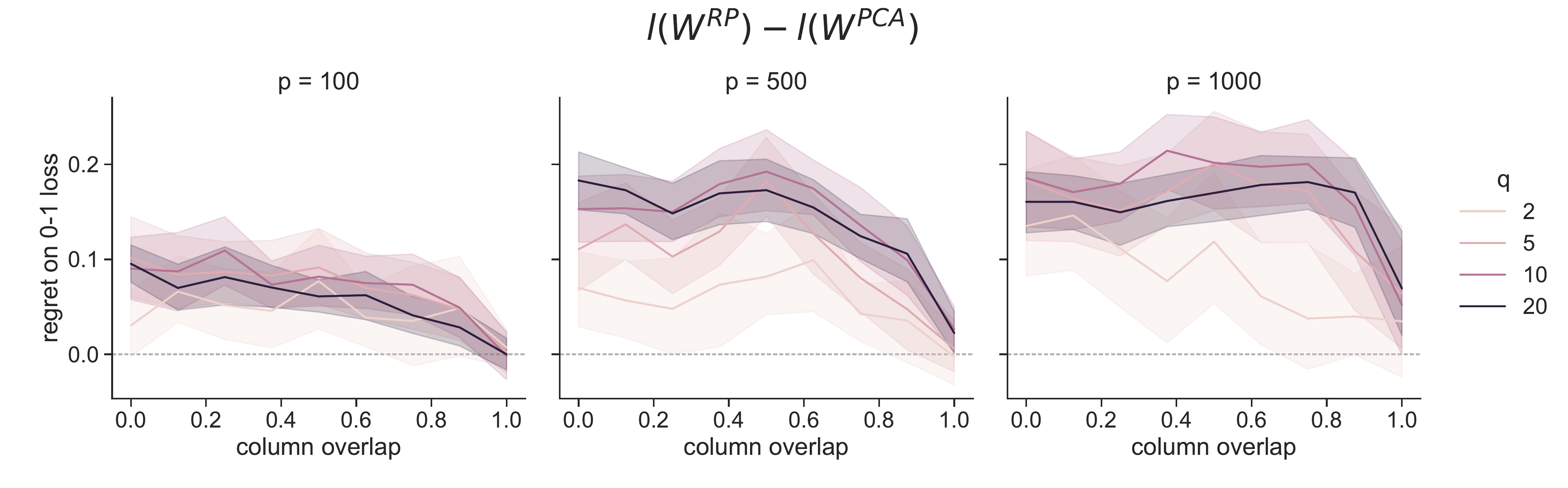}}\\
    \subfloat{\includegraphics[width=\linewidth]{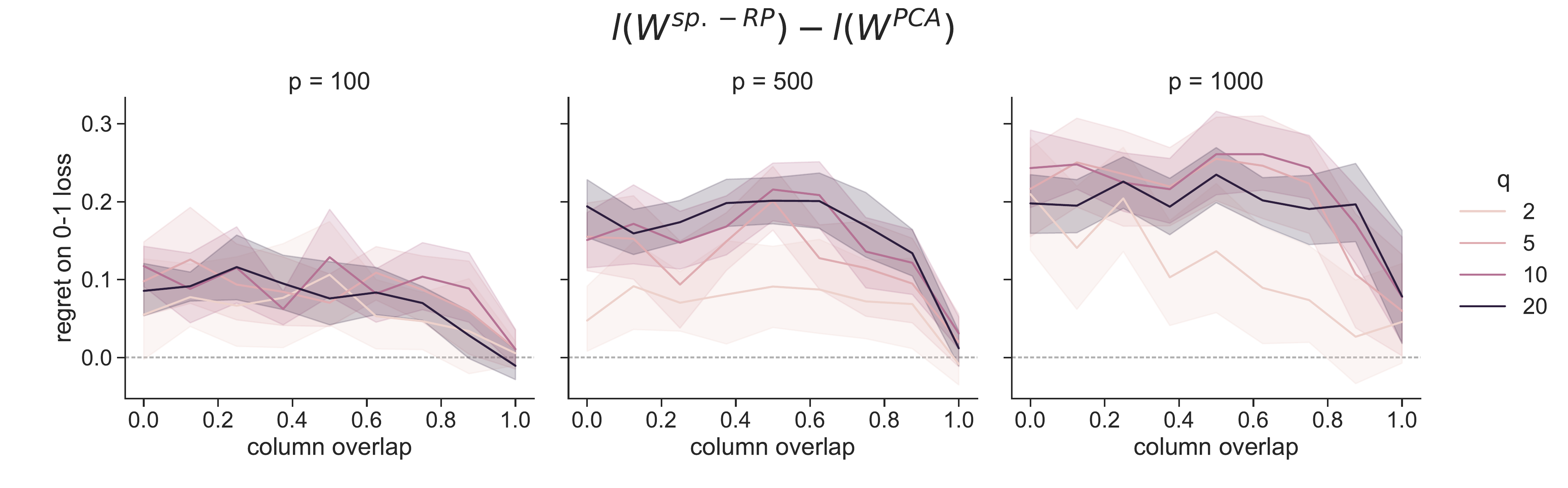}}\\
    \caption{Regret of training a classifier within a RP (up) or sparse-RP (down) subspace as opposed to a PCA subspace. Comparatively, PCA projection seems to be at its best for larger values of $p$. When the column overlap $\gamma$ grows to 1, the two classes are fundamentally the same, the error with any projection is close to 0.5, and the regret close to 0.
    }
    \label{fig:finite_data_regret}
\end{figure*}

For a more detailed exploration of the impact of the dimension settings $p$ and $q$, we display on \figurename~\ref{fig:finite_data_regret} the regret of $l(W^{\text{PCA}}) - l(W^{\text{RP}})$ of choosing PCA projection over RP against the overlap $\gamma$. In particular, we can note that PCA seems particularly preferable in high dimension $p$.

\section{Comparison with the Bhattacharyya-optimal projection}\label{sect:comparison_optimal}
In this section, we take a closer look at how the Bhattacharyya-optimal projection, defined in Section \ref{sect:optimal_subspace} performs on both the formal infinite Gaussian data scenario and on more concrete finite data cases. This projection is constructed with full knowledge of the labels, hence would not be applicable in the settings of interest to us, but provides nevertheless an interesting  benchmark.

\begin{figure*}[tbhp]
    \centering
    \includegraphics[width=\linewidth]{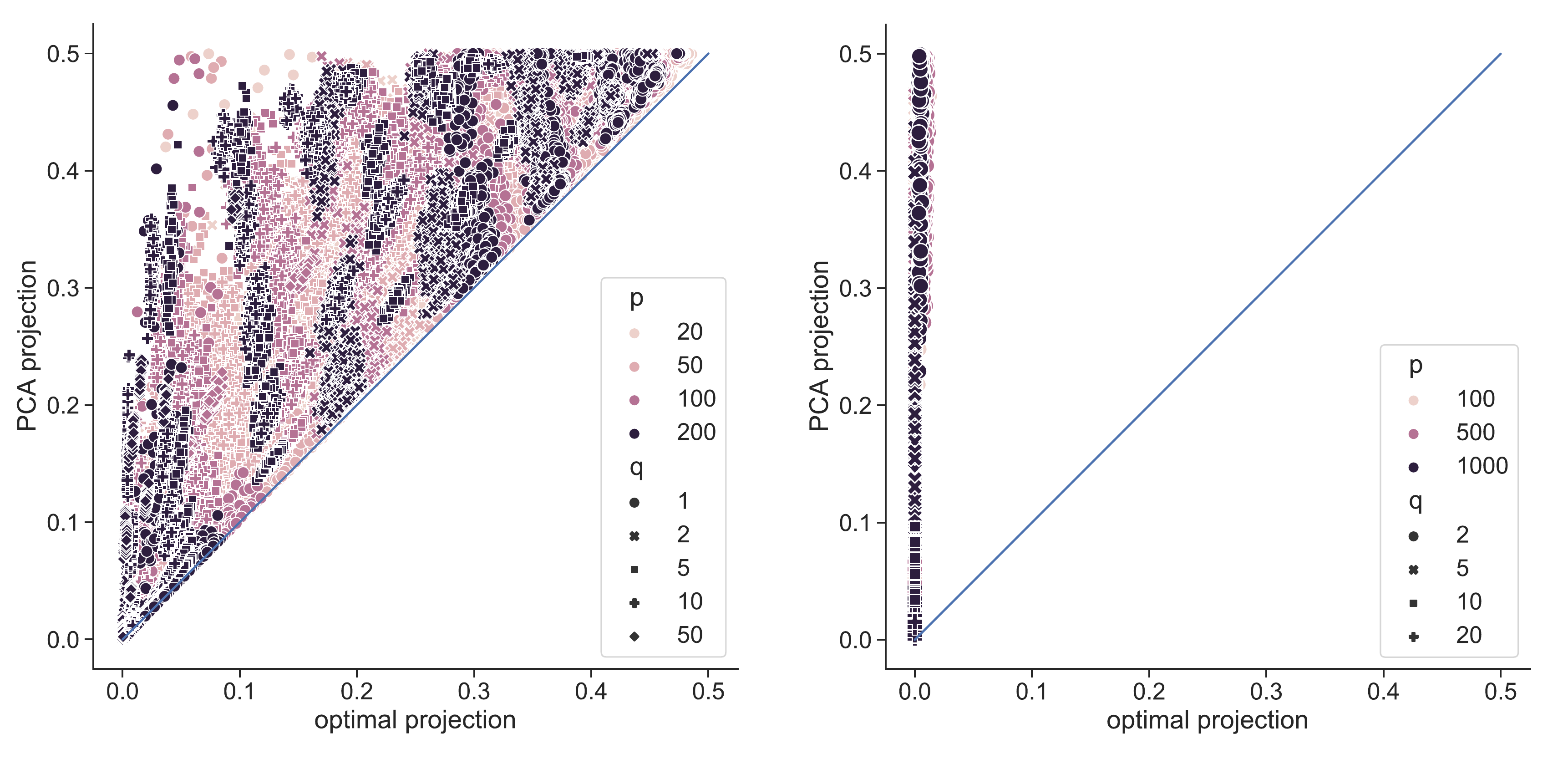}
    \caption{Bhattacharyya-overlap within the embedding subspace. PCA ($y$-axis) vs optimal projection ($x$-axes). (Left) 264600 pairs of covariance matrices generated as inverse Wishart with several different pairs of degrees of freedom, as in \figurename~\ref{fig:scatterplot_IW}. (Right) 6480 pairs covariance matrices generated as the empirical covariance matrices of real RNA sequencing data, with several levels of overlap between the columns of the real datasets, as in \figurename~\ref{fig:scatterplot_RNA_seq}.}
    \label{fig:PCA_vs_Optimal}
\end{figure*}

First, we analyse the ideal theoretical version of the Bhattacharyya-optimal projection, defined from the true covariance matrices $\Sigma_1$ and $\Sigma_2$. In \figurename~\ref{fig:PCA_vs_Optimal} we compare the Bhattacharyya-overlap, under Gaussian data assumption, of the optimal and PCA projections for two different families of covariance matrices. On the left, the two covariance matrices are generated as inverse Wishart, with the same experimental settings as in \figurename~\ref{fig:scatterplot_IW}. On the right, they are generated as the empirical covariances of real data, with the same settings \figurename~\ref{fig:scatterplot_RNA_seq}. As expected, the Bhattacharyya-optimal projection dominates in all cases.

\begin{figure*}[tbhp]
    \centering
    \includegraphics[width=\linewidth]{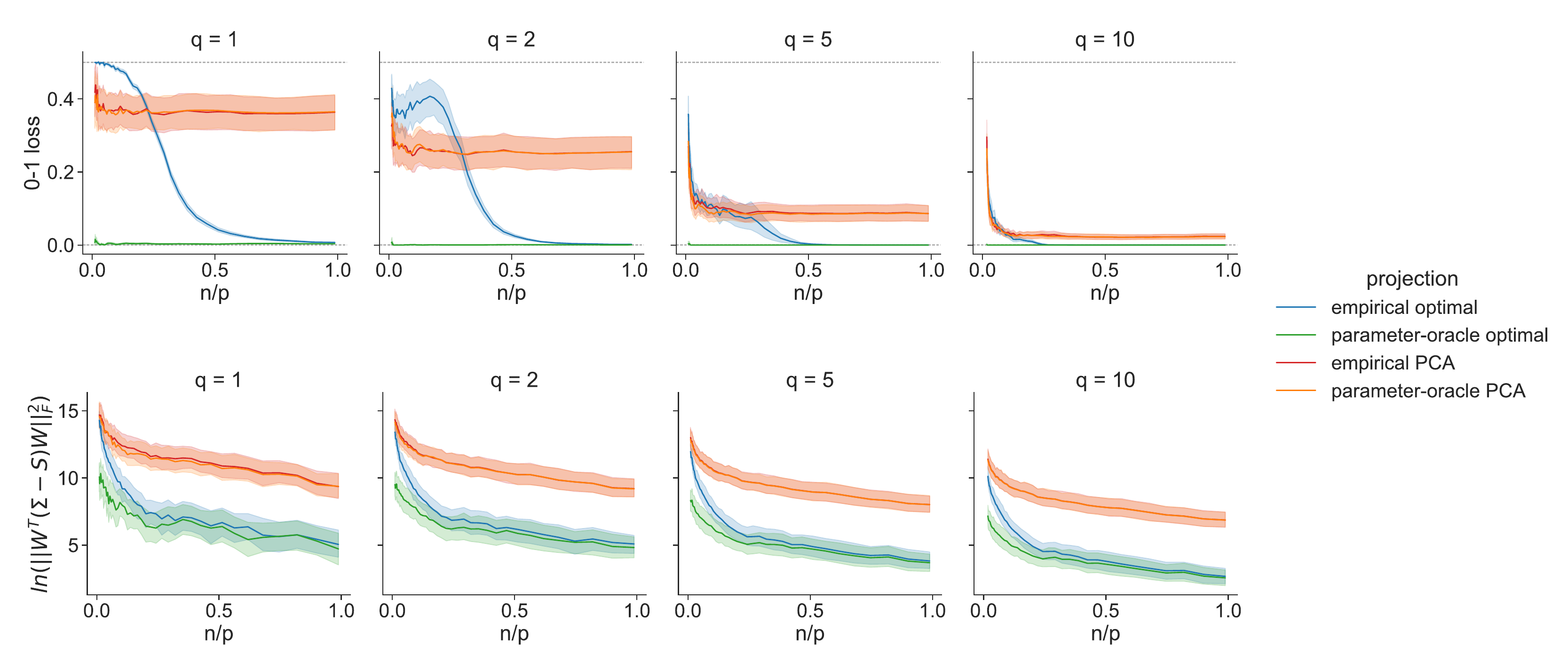}
    \caption{(Top) Out of Sample 0-1 loss against total sample size of likelihood-ratio classifiers trained in the embedding subspaces defined by different projections. (Bottom) Logarithm of the reconstruction error on the covariance $\Sigma_k$ in the embedding spaces. For all figures the data is Gaussian, with ground truth covariance matrices $\Sigma_k$ generated from real RNA sequencing data. The ambient space is of size $p=1000$. There are two balanced classes in the data, with sample size $n$ each, for a total sample size of $2n$. As per the procedure described in \figurename~\ref{fig:column_overlap}, a distribution overlap of $\gamma = 0.2$ was enforced between the columns of the real data before generating the covariance matrices. In terms of 0-1 loss, the empirical version of the Bhattacharyya-optimal projection does not outperform PCA when the sample size is small. Additionally, the gap in performances between the \textit{parameter-oracle} and \textit{empirical} versions is much wider for the optimal than for PCA.}
    \label{fig:PCA_vs_Optimal_finite_sample}
\end{figure*}

However, with finite sample size, the true matrices $\Sigma_1$ and $\Sigma_2$ are unavailable: only their empirical estimates $S_1$ and $S_2$ can be used to learn the projection. This is crucial, since the matrix inversion involved in the definition of the Bhattacharyya-optimal projection can magnify even small deviations between a covariance matrix $\Sigma$ and its empirical estimate $S$. In the following, we demonstrate that the projection defined by plugging $S_1$ and $S_2$ in the Bhattacharyya-optimal formula, which we call the \textit{empirical} Bhattacharyya-optimal projection, does not outperform the \textit{empirical} PCA as dramatically as its \textit{parameter-oracle} counterpart does. For the experimental design, we re-use the matrix generation scheme of \figurename~\ref{fig:scatterplot_RNA_seq}. However, after two covariance matrices $(\Sigma_1, \Sigma_2)$ are generated, instead of computing the corresponding Bhattacharyya-overlap with the projection of interest, we generate finite Gaussian data from the matrices, $X_k \sim \N(0_p, \Sigma_k)$, learn a projection from a training subset of this data, embed the data accordingly, train a likelihood ratio classifier with the embedded training set, and evaluate the performances of this classifier the remaining validation data. The training set is made of $70\%$ of the total dataset, the validation set of the remaining $30\%$. We compare the Bhattacharyya-optimal and PCA projections within this experimental setting. Each projection has a \textit{parameter-oracle} version, a virtual ideal scenario where it is computed with the knowledge of the true covariance matrices $\Sigma_1$ and $\Sigma_2$, and an \textit{empirical} version, a more realistic scenario where it is instead computed from the empirical covariances $S_1$ and $S_2$, estimated on the training set. The Out of Sample 0-1 loss of the classifiers are displayed on the first row of \figurename~\ref{fig:PCA_vs_Optimal_finite_sample} as a function of the total sample size $n$. Although the \textit{parameter-oracle} optimal projection performs very well, its \textit{empirical} version is significantly worse, no longer outperforming PCA. On the other hand, the PCA projection is strikingly stable in the finite sample case, with its \textit{parameter-oracle} and \textit{empirical} version performing very similarly. This behaviour is due to the fact that the \textit{empirical} optimal projection selects the eigenspaces of $S_1^{-1} S_2$ with the highest and lowest eigenvalues possible, whereas \textit{empirical} PCA selects the eigenspaces of $S_1 {+} S_2$ with the highest eigenvalues. With finite data, the latter is much more stable and the selected eigenspaces have a better chance of being close to the oracle eigenspaces of $\Sigma_1 {+} \Sigma_2$. On the other hand, the extreme eigenspaces of $S_1^{-1} S_2$ can be degenerate, since they rely on the {\it smallest} eigenvalues of matrices which are either approximately low-rank or truly low-rank. When the matrices are truly low-rank, regularisation is actually necessary to approximate the inversion.

To gain further insight into these phenomena, in the second row of \figurename~\ref{fig:PCA_vs_Optimal_finite_sample} we show (log) reconstruction error $\frac{1}{2}\sum_{k=1}^2 \norm{W^t(S_k-\Sigma_k)W}_F^2$ for the covariance matrices in the respective subspace. Reconstruction error with the \textit{empirical} Bhattacharyya-optimal projection improves dramatically with sample size. 
The level of error starts similar (although smaller) to the PCA errors, and ends similar (although higher) to the \textit{parameter-oracle} error. 
However, as already seen in the first row of \figurename~\ref{fig:PCA_vs_Optimal_finite_sample}) 
these improvements do not lead to better 0-1 performance. For example, for sample sizes below $0.4 p$, the 0-1 loss with the \textit{empirical} Bhattacharyya-optimal projection remains fairly similar to the 0-1 loss with the PCA projections, despite the matrix reconstruction error of PCA being orders of magnitude worse. This suggest that the reason for relatively poor performance wrt 0-1 loss under the \textit{empirical} Bhattacharyya-optimal projection (relative to its \textit{parameter-oracle} version) may be more due to poor subspace selection than to poor matrix estimation within the subspace. Which in turn supports the idea that the subspace selection of the Bhattacharyya-optimal procedure is impacted in finite sample size scenarios.

\begin{figure*}[tbhp]
    \centering
    \includegraphics[width=\linewidth]{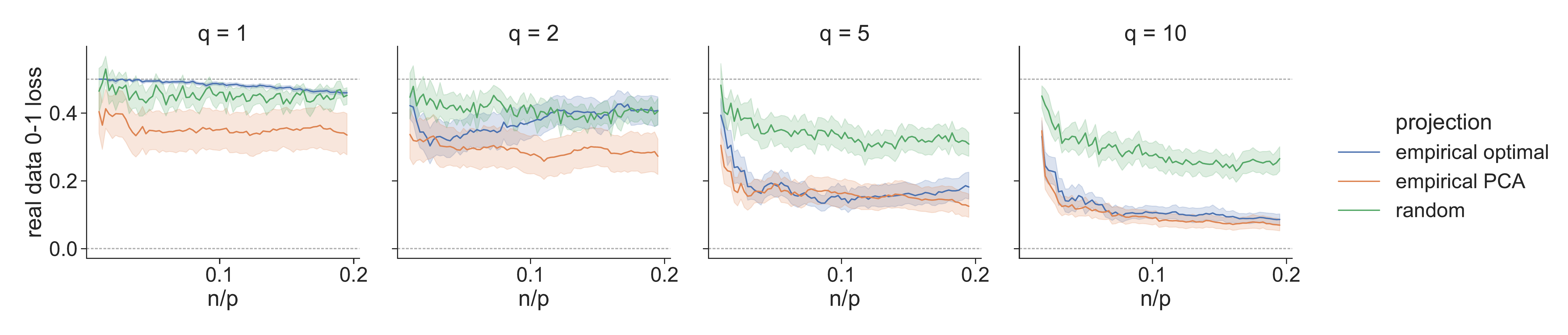}
    \caption{Out of Sample 0-1 loss against total sample size of likelihood-ratio classifiers trained in the embedding subspaces defined by different projections. The data is real RNA sequencing, with no ground truth for the covariance matrices. The ambient space is of size $p=1000$. There are two balanced classes in the data, with sample size $n \leq 195$ each, for a total sample size of $2n \leq 390$. As per the procedure described in \figurename~\ref{fig:column_overlap}, a distribution overlap of $\gamma = 0.2$ was enforced between the columns of the real data. The Bhattacharyya-optimal projection is not dominant on real data with finite sample size.}
    \label{fig:optimal_vs_pca_real_data_error}
\end{figure*}

We can expand even further on these results by relaxing the Gaussian assumption of the data. In \figurename~\ref{fig:optimal_vs_pca_real_data_error}, we evaluate the projections and their associated embedded classifier on the RNA sequencing dataset. The \textit{parameter-oracle} projections cannot be defined anymore in this setting: all projections are \textit{empirical}. RP  is also included for comparison. Here we see that on real finite data the \textit{empirical} Bhattacharyya-optimal projection does not outperform PCA (or even RP in some cases) as much as its \textit{parameter-oracle} version does on Gaussian data, both infinite and finite.

\section{Discussion and conclusion}\label{sect:discussion}

\begin{figure*}[tbhp]
    \centering
    \includegraphics[width=\linewidth]{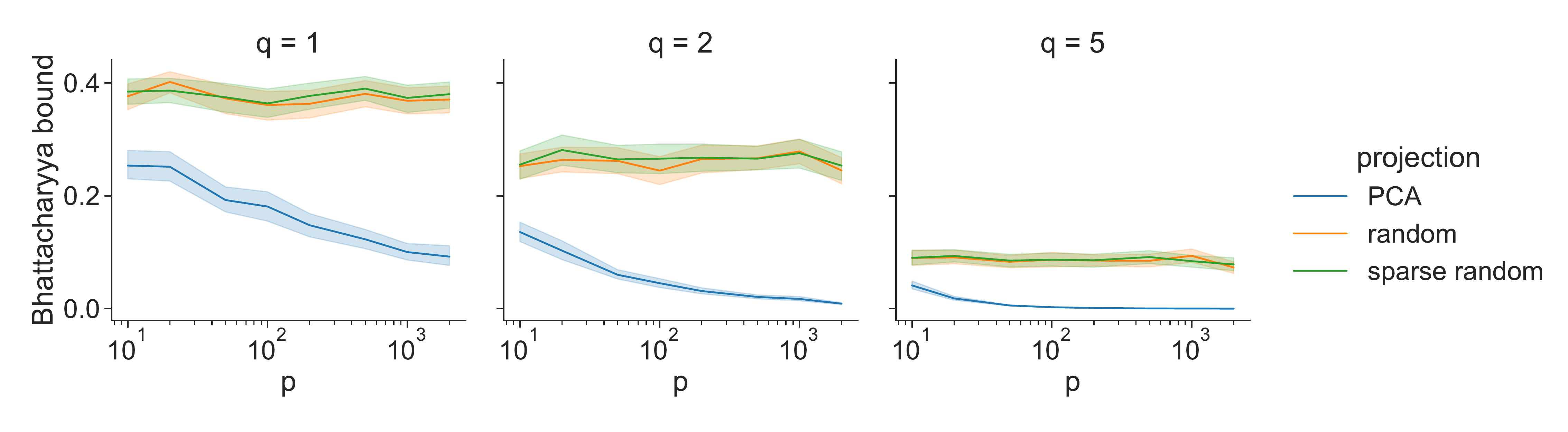}
    \caption{Bhattacharyya-overlap within an embedding space of size $q$ as a function of the ambient size $p$, displayed in logarithmic scale. The covariance matrices follow the inverse Wishart distribution. In the case of the PCA projection, we observe the effects of the "blessing of dimensionality". The performances improve quite quickly with $p$. Confidence intervals $95\%$, 100 simulations for each combination of the settings.}
    \label{fig:relplot_p}
\end{figure*}

{\it The behaviour of PCA in high dimensions.}
We saw that in a number of settings PCA performs well in terms of retaining relevant information. A fuller theoretical understanding of this behaviour is beyond the scope of this paper, but we offer some initial comments here.
As discussed above, 
PCA on the full data (i.e. spanning the classes, but without labels) selects the highest eigenvalues of $\Sigma_1 {+} \Sigma_2$ (for simplicity of argument, here we assume balanced classes). Hence, the factor that determines the performance of PCA projection is similarity of the respective principal components of the individual matrices $\Sigma_1$ and $\Sigma_2$. When there exist many directions in which both matrices are strong with similar amplitude, PCA projection will be poor. This is the case of example 2 of \sectionname~\ref{sect:PCA_theory}. On the other hand, when these directions are not too similar, PCA will perform better, 
as in example 1 of \sectionname~\ref{sect:PCA_theory}. 
As far as we know, detailed behaviour in this context has not been fully studied theoretically. 
An interesting notion is that as the dimension $p$ increases, 
alignment of the first few principal components of both classes may become increasingly unlikely, leading to a potential 
``blessing of dimensionality''.
For instance, the eigenvector matrices of Wishart matrices with identity scale follow the Haar distribution, that is to say uniform over $O_p(\R)$ \citep{bai2007asymptotics}. As a result, 
in high dimensions
it is unlikely for two independent Wishart matrices 
to have similar eigenvectors. \figurename~\ref{fig:relplot_p} shows a ``blessing of dimensionality'' of this kind for PCA projection. We observe that the Bhattacharyya-overlap {\it decreases} with $p$ for PCA, but not for RP.

\medskip
\noindent
{\it Conclusions}. 
We studied the behaviour of linear projections for the purpose of retaining class-specific second order signals. Our work was  motivated by unsupervised settings in which no prior knowledge of the class labels can be used for projection. In particular, we 
sought to understand whether PCA applied to full mixed data (i.e. spanning the classes but without labels) could preserve 
discriminative differences in covariance structure between classes. While without restriction on  covariance structures there can be no fully general guarantee in this regard, via a quasi-exhaustive enumeration protocol we were able to study behaviour over a very wide range of covariance structures. We also studied real data examples. Taken together, the results demonstrate that PCA-defined subspaces are very often effectively able to retain covariance signals
and are, in particular, more effective than those obtained by  random projection. Comparison between the finite and infinite data versions of the PCA and Bhattacharyya-optimal projections revealed that PCA  is  much more robust to the sample size.
In addition, PCA projection can actually benefit from the ambient space size increasing. Although further theoretical work is needed, it seems there may be a``blessing of dimensionality" effect, wherein the strong eigenvectors of the mixed data are more likely to differ between classes in higher dimensions, since pairs of class-specific eigenvectors have a lower chance to be collinear.
To conclude, our results suggest that PCA can be usefully employed as a step in high-dimensional unsupervised learning settings where signals may lie at the level of differential covariance structure. 


\paragraph{Acknowledgments}
The authors would like to thank Ata Kab\'{a}n for her valuable comments which were very helpful to improve this work. Supported by the 
Bundesministerium f\"{u}r Bildung und Forschung (BMBF) project ``MechML".

\bibliography{references}

\begin{thebibliography}{}

\bibitem[Achlioptas, 2003]{achlioptas2003database}
Achlioptas, D. (2003).
\newblock {Database-friendly random projections: Johnson-Lindenstrauss with
  binary coins}.
\newblock {\em Journal of computer and System Sciences}, 66(4):671--687.

\bibitem[Achlioptas and McSherry, 2005]{achlioptas2005spectral}
Achlioptas, D. and McSherry, F. (2005).
\newblock On spectral learning of mixtures of distributions.
\newblock In {\em International Conference on Computational Learning Theory},
  pages 458--469. Springer.

\bibitem[Ailon and Chazelle, 2009]{ailon2009fast}
Ailon, N. and Chazelle, B. (2009).
\newblock {The fast Johnson--Lindenstrauss transform and approximate nearest
  neighbors}.
\newblock {\em SIAM Journal on computing}, 39(1):302--322.

\bibitem[Bai et~al., 2007]{bai2007asymptotics}
Bai, Z., Miao, B., and Pan, G. (2007).
\newblock On asymptotics of eigenvectors of large sample covariance matrix.
\newblock {\em The Annals of Probability}, 35(4):1532--1572.

\bibitem[Bhattacharyya, 1943]{bhattacharyya1943measure}
Bhattacharyya, A. (1943).
\newblock On a measure of divergence between two statistical populations
  defined by their probability distributions.
\newblock {\em Bull. Calcutta Math. Soc.}, 35:99--109.

\bibitem[Brubaker and Vempala, 2008]{brubaker2008isotropic}
Brubaker, S.~C. and Vempala, S.~S. (2008).
\newblock {Isotropic PCA and affine-invariant clustering}.
\newblock In {\em Building Bridges}, pages 241--281. Springer.

\bibitem[Cannings and Samworth, 2017]{cannings2017random}
Cannings, T.~I. and Samworth, R.~J. (2017).
\newblock Random-projection ensemble classification.
\newblock {\em Journal of the Royal Statistical Society: Series B (Statistical
  Methodology)}, 79(4):959--1035.

\bibitem[Chernoff, 1952]{chernoff1952measure}
Chernoff, H. (1952).
\newblock A measure of asymptotic efficiency for tests of a hypothesis based on
  the sum of observations.
\newblock {\em The Annals of Mathematical Statistics}, pages 493--507.

\bibitem[Choi and Lee, 2003]{choi2003feature}
Choi, E. and Lee, C. (2003).
\newblock {Feature extraction based on the Bhattacharyya distance}.
\newblock {\em Pattern Recognition}, 36(8):1703--1709.

\bibitem[Dasgupta, 1999]{dasgupta1999learning}
Dasgupta, S. (1999).
\newblock Learning mixtures of {G}aussians.
\newblock In {\em 40th Annual Symposium on Foundations of Computer Science
  (Cat. No. 99CB37039)}, pages 634--644. IEEE.

\bibitem[Dasgupta and Gupta, 2003]{dasgupta2003elementary}
Dasgupta, S. and Gupta, A. (2003).
\newblock An elementary proof of a theorem of {J}ohnson and {L}indenstrauss.
\newblock {\em Random Structures \& Algorithms}, 22(1):60--65.

\bibitem[Dasgupta and Schulman, 2000]{dasgupta2000two}
Dasgupta, S. and Schulman, L.~J. (2000).
\newblock {A Two-Round Variant of {EM} for Gaussian Mixtures}.
\newblock In {\em Proceedings of the 16th Conference on Uncertainty in
  Artificial Intelligence}, UAI '00, page 152–159, San Francisco, CA, USA.
  Morgan Kaufmann Publishers Inc.

\bibitem[Deegalla and Bostrom, 2006]{deegalla2006reducing}
Deegalla, S. and Bostrom, H. (2006).
\newblock Reducing high-dimensional data by principal component analysis vs.
  random projection for nearest neighbor classification.
\newblock In {\em 2006 5th International Conference on Machine Learning and
  Applications (ICMLA'06)}, pages 245--250. IEEE.

\bibitem[Ding and He, 2004]{ding2004k}
Ding, C. and He, X. (2004).
\newblock K-means clustering via principal component analysis.
\newblock In {\em Proceedings of the twenty-first international conference on
  Machine learning}, page~29.

\bibitem[Durrant and Kab{\'a}n, 2010]{durrant2010compressed}
Durrant, R. and Kab{\'a}n, A. (2010).
\newblock {Compressed Fisher linear discriminant analysis: Classification of
  randomly projected data}.
\newblock In {\em Proceedings of the 16th ACM SIGKDD international conference
  on Knowledge discovery and data mining}, pages 1119--1128.

\bibitem[Durrant and Kab{\'a}n, 2013]{durrant2013sharp}
Durrant, R. and Kab{\'a}n, A. (2013).
\newblock {Sharp generalization error bounds for randomly-projected
  classifiers}.
\newblock In {\em International Conference on Machine Learning}, pages
  693--701. PMLR.

\bibitem[Durrant and Kab{\'a}n, 2015]{durrant2015random}
Durrant, R. and Kab{\'a}n, A. (2015).
\newblock Random projections as regularizers: learning a linear discriminant
  from fewer observations than dimensions.
\newblock {\em Machine Learning}, 99(2):257--286.

\bibitem[Edelman, 1991]{edelman1991distribution}
Edelman, A. (1991).
\newblock {The distribution and moments of the smallest eigenvalue of a random
  matrix of Wishart type}.
\newblock {\em Linear algebra and its applications}, 159:55--80.

\bibitem[Everitt and Hand, 1981]{everitt1981finite}
Everitt, B. and Hand, D. (1981).
\newblock Finite mixture distributions.
\newblock {\em Monographs on Applied Probability and Statistics}.

\bibitem[Feldman et~al., 2006]{feldman2006pac}
Feldman, J., Servedio, R.~A., and O’Donnell, R. (2006).
\newblock {PAC learning axis-aligned mixtures of Gaussians with no separation
  assumption}.
\newblock In {\em International Conference on Computational Learning Theory},
  pages 20--34. Springer.

\bibitem[Fisher, 1936]{fisher1936use}
Fisher, R.~A. (1936).
\newblock The use of multiple measurements in taxonomic problems.
\newblock {\em Annals of eugenics}, 7(2):179--188.

\bibitem[Fradkin and Madigan, 2003]{fradkin2003experiments}
Fradkin, D. and Madigan, D. (2003).
\newblock Experiments with random projections for machine learning.
\newblock In {\em Proceedings of the ninth ACM SIGKDD international conference
  on Knowledge discovery and data mining}, pages 517--522.

\bibitem[Fukunaga, 1972]{fukunaga1972instruction}
Fukunaga, K. (1972).
\newblock {\em {Instruction to Statistical Pattern Recognition}}.
\newblock Elsevier.

\bibitem[Guorong et~al., 1996]{guorong1996bhattacharyya}
Guorong, X., Peiqi, C., and Minhui, W. (1996).
\newblock Bhattacharyya distance feature selection.
\newblock In {\em Proceedings of 13th International Conference on Pattern
  Recognition}, volume~2, pages 195--199. IEEE.

\bibitem[Hastie and Tibshirani, 1996]{hastie1996discriminant}
Hastie, T. and Tibshirani, R. (1996).
\newblock {Discriminant analysis by Gaussian mixtures}.
\newblock {\em Journal of the Royal Statistical Society: Series B
  (Methodological)}, 58(1):155--176.

\bibitem[Henderson and Lainiotis, 1969]{henderson1969comments}
Henderson, T. and Lainiotis, D. (1969).
\newblock Comments on linear feature extraction [{C}orresp.].
\newblock {\em IEEE Transactions on Information Theory}, 15(6):728--730.

\bibitem[Hertrich et~al., 2020]{hertrich2020pca}
Hertrich, J., Nguyen, D. P.~L., Aujol, J.-F., Bernard, D., Berthoumieu, Y.,
  Saadaldin, A., and Steidl, G. (2020).
\newblock {PCA Reduced Gaussian Mixture Models with Applications in
  Superresolution}.
\newblock {\em arXiv preprint arXiv:2009.07520}.

\bibitem[Houdard et~al., 2018]{houdard2018high}
Houdard, A., Bouveyron, C., and Delon, J. (2018).
\newblock {High-dimensional mixture models for unsupervised image denoising
  (HDMI)}.
\newblock {\em SIAM Journal on Imaging Sciences}, 11(4):2815--2846.

\bibitem[Ideker and Krogan, 2012]{ideker2012differential}
Ideker, T. and Krogan, N.~J. (2012).
\newblock Differential network biology.
\newblock {\em Molecular systems biology}, 8(1):565.

\bibitem[Jin and Wang, 2016]{jin2016influential}
Jin, J. and Wang, W. (2016).
\newblock Influential features {PCA} for high dimensional clustering.
\newblock {\em The Annals of Statistics}, 44(6):2323--2359.

\bibitem[Kab{\'a}n, 2020]{kaban2020sufficient}
Kab{\'a}n, A. (2020).
\newblock Sufficient ensemble size for random matrix theory-based handling of
  singular covariance matrices.
\newblock {\em Analysis and Applications}, 18(05):929--950.

\bibitem[Kab{\'a}n and Durrant, 2020]{kaban2020structure}
Kab{\'a}n, A. and Durrant, R. (2020).
\newblock {Structure from Randomness in Halfspace Learning with the Zero-One
  Loss}.
\newblock {\em Journal of Artificial Intelligence Research}, 69:733--764.

\bibitem[Kadota and Shepp, 1967]{kadota1967best}
Kadota, T. and Shepp, L. (1967).
\newblock {On the best finite set of linear observables for discriminating two
  Gaussian signals}.
\newblock {\em IEEE Transactions on Information Theory}, 13(2):278--284.

\bibitem[Kannan et~al., 2005]{kannan2005spectral}
Kannan, R., Salmasian, H., and Vempala, S. (2005).
\newblock The spectral method for general mixture models.
\newblock In {\em International Conference on Computational Learning Theory},
  pages 444--457. Springer.

\bibitem[Kim et~al., 2003]{kim2003efficient}
Kim, H.-C., Kim, D., and Bang, S.~Y. (2003).
\newblock {An efficient model order selection for PCA mixture model}.
\newblock {\em Pattern Recognition Letters}, 24(9-10):1385--1393.

\bibitem[Kullback, 1959]{kullback1997information}
Kullback, S. (1997 [1959]).
\newblock {\em Information theory and statistics}.
\newblock Courier Corporation.

\bibitem[Li et~al., 2006]{li2006very}
Li, P., Hastie, T.~J., and Church, K.~W. (2006).
\newblock Very sparse random projections.
\newblock In {\em Proceedings of the 12th ACM SIGKDD international conference
  on Knowledge discovery and data mining}, pages 287--296.

\bibitem[Lindenstrauss and Johnson, 1984]{lindenstrauss1984extensions}
Lindenstrauss, W. and Johnson, J. (1984).
\newblock Extensions of {L}ipschitz maps into a {H}ilbert space.
\newblock {\em Contemp. Math}, 26:189--206.

\bibitem[Mika et~al., 1999]{mika1999fisher}
Mika, S., Ratsch, G., Weston, J., Scholkopf, B., and Mullers, K.-R. (1999).
\newblock Fisher discriminant analysis with kernels.
\newblock In {\em Neural networks for signal processing IX: Proceedings of the
  1999 IEEE signal processing society workshop (cat. no. 98th8468)}, pages
  41--48. Ieee.

\bibitem[Pickles, 1985]{pickles1985introduction}
Pickles, A. (1985).
\newblock {\em An introduction to likelihood analysis}.
\newblock Number~42 in Concepts and techniques in modern geography. Geo Books
  Norwich, UK.

\bibitem[Rao and Varadarajan, 1963]{rao1963discrimination}
Rao, C.~R. and Varadarajan, V. (1963).
\newblock {Discrimination of Gaussian processes}.
\newblock {\em Sankhy{\=a}: The Indian Journal of Statistics, Series A}, pages
  303--330.

\bibitem[Reboredo et~al., 2014]{reboredo2014compressive}
Reboredo, H., Renna, F., Calderbank, R., and Rodrigues, M.~R. (2014).
\newblock {Compressive classification of a mixture of Gaussians: Analysis,
  designs and geometrical interpretation}.
\newblock {\em arXiv preprint arXiv:1401.6962}.

\bibitem[Reeve and Kaban, 2021]{reeve2021statistical}
Reeve, H.~W. and Kaban, A. (2021).
\newblock Statistical optimality conditions for compressive ensembles.
\newblock {\em arXiv preprint arXiv:2106.01092}.

\bibitem[Sanjeev and Kannan, 2001]{sanjeev2001learning}
Sanjeev, A. and Kannan, R. (2001).
\newblock {Learning Mixtures of Arbitrary Gaussians}.
\newblock In {\em Proceedings of the thirty-third annual ACM symposium on
  Theory of computing}, pages 247--257.

\bibitem[Scrucca et~al., 2016]{scrucca2016mclust}
Scrucca, L., Fop, M., Murphy, T.~B., and Raftery, A.~E. (2016).
\newblock mclust 5: clustering, classification and density estimation using
  {G}aussian finite mixture models.
\newblock {\em The R journal}, 8(1):289.

\bibitem[Severini, 2000]{severini2000likelihood}
Severini, T.~A. (2000).
\newblock {\em Likelihood methods in statistics}.
\newblock Oxford University Press.

\bibitem[Smith et~al., 2011]{smith2011network}
Smith, S.~M., Miller, K.~L., Salimi-Khorshidi, G., Webster, M., Beckmann,
  C.~F., Nichols, T.~E., Ramsey, J.~D., and Woolrich, M.~W. (2011).
\newblock Network modelling methods for {FMRI}.
\newblock {\em Neuroimage}, 54(2):875--891.

\bibitem[St{\"a}dler et~al., 2017]{stadler2017molecular}
St{\"a}dler, N., Dondelinger, F., Hill, S.~M., Akbani, R., Lu, Y., Mills,
  G.~B., and Mukherjee, S. (2017).
\newblock Molecular heterogeneity at the network level: high-dimensional
  testing, clustering and a {TCGA} case study.
\newblock {\em Bioinformatics}, 33(18):2890--2896.

\bibitem[Sugiyama, 2007]{sugiyama2007dimensionality}
Sugiyama, M. (2007).
\newblock Dimensionality reduction of multimodal labeled data by local {F}isher
  discriminant analysis.
\newblock {\em Journal of machine learning research}, 8(5).

\bibitem[Taschler et~al., 2019]{taschler2019model}
Taschler, B., Dondelinger, F., and Mukherjee, S. (2019).
\newblock Model-based clustering in very high dimensions via adaptive
  projections.
\newblock {\em arXiv preprint arXiv:1902.08472}.

\bibitem[Van~Trees, 1968]{van2004detection}
Van~Trees, H.~L. (2004 [1968]).
\newblock {\em Detection, estimation, and modulation theory, part I: detection,
  estimation, and linear modulation theory}.
\newblock John Wiley \& Sons.

\bibitem[Varoquaux et~al., 2010]{varoquaux2010brain}
Varoquaux, G., Gramfort, A., Poline, J.~B., and Thirion, B. (2010).
\newblock Brain covariance selection: better individual functional connectivity
  models using population prior.
\newblock In {\em Proceedings of {NeurIPS} 2010}.

\bibitem[Vempala and Wang, 2002]{vempala2002spectral}
Vempala, S. and Wang, G. (2002).
\newblock A spectral algorithm for learning mixtures of distributions.
\newblock In {\em The 43rd Annual IEEE Symposium on Foundations of Computer
  Science, 2002. Proceedings.}, pages 113--122. IEEE.

\bibitem[Xu et~al., 2014]{xu2014multimode}
Xu, X., Xie, L., and Wang, S. (2014).
\newblock {Multimode process monitoring with PCA mixture model}.
\newblock {\em Computers \& Electrical Engineering}, 40(7):2101--2112.

\bibitem[Xuan et~al., 2006]{xuan2006feature}
Xuan, G., Zhu, X., Chai, P., Zhang, Z., Shi, Y.~Q., and Fu, D. (2006).
\newblock {Feature selection based on the Bhattacharyya distance}.
\newblock In {\em 18th International Conference on Pattern Recognition
  (ICPR'06)}, volume~4, pages 957--957. IEEE.

\bibitem[Zhang and Kab{\'a}n, 2019]{zhang2019experiments}
Zhang, X. and Kab{\'a}n, A. (2019).
\newblock {Experiments with Random Projections Ensembles: Linear Versus
  Quadratic Discriminants.}
\newblock In {\em ICDM Workshops}, pages 719--726.

\end{thebibliography}

\end{document}